\pdfoutput=1

\documentclass[11pt]{article}

\usepackage[]{acl}

\usepackage{booktabs} %
\usepackage{arydshln}
\usepackage{stfloats}
\usepackage{amsmath}
\usepackage{balance}
\usepackage{graphicx}
\usepackage{bbm}
\usepackage{amssymb}
\usepackage{color,xcolor}
\usepackage{geometry}
\usepackage{mathptmx}
\usepackage{bm}
\usepackage{tikz}
\usepackage{caption, subcaption}
\usepackage{colortbl}
\usepackage{adjustbox}
\usepackage{float}
\usepackage{pgfplots}
\usepackage{microtype}
\usepackage{tabularx}
\usepackage{graphicx}
\usepackage{float}
\usepackage{makecell}
\usepackage{booktabs}
\usepackage{color}
\usepackage{xargs}
\usepackage[colorinlistoftodos,prependcaption,textsize=tiny]{todonotes}
\usepackage{times}
\usepackage{helvet}
\usepackage{courier}
\usepackage{amsmath, xparse}
\usepackage{amssymb}
\usepackage{graphicx}
\usepackage{natbib}
\usepackage{multirow}
\usepackage{booktabs}
\usepackage{float}
\usepackage{soul}
\usepackage{algorithm,algpseudocode}

\pgfplotsset{
axis background/.style={fill=gallery},
grid=both,
  xtick pos=left,
  ytick pos=left,
  y label style={at={(0.03,0.5)}},
}
\definecolor{tea_green}{RGB}{214, 234, 193}
\definecolor{hint_green}{RGB}{226,246,209}
\definecolor{Madang}{RGB}{190,235,159}
\definecolor{yellow_green}{RGB}{198,222,119}
\definecolor{link_water}{RGB}{221, 232, 250}
\definecolor{celestial_blue}{RGB}{52, 152, 219}
\definecolor{shakespeare}{RGB}{85, 154, 193}
\definecolor{buttermilk}{RGB}{255,242,174}
\definecolor{chardonnay}{RGB}{250,196,114}
\definecolor{rajah}{RGB}{253,180,98}
\definecolor{fog}{RGB}{213, 193, 234}
\definecolor{melon}{RGB}{254,191,181}
\definecolor{sundown}{RGB}{249, 180, 181}
\definecolor{mona_lisa}{RGB}{246,152,134}
\definecolor{salmon}{RGB}{242,131,107}

\definecolor{saltpan}{RGB}{238, 243, 232}
\definecolor{aqua_spring}{RGB}{232, 243, 232}
\definecolor{tea_green}{RGB}{214, 234, 193}
\definecolor{Madang}{RGB}{190,235,159}
\definecolor{fringy_flower}{RGB}{194, 234, 193}
\definecolor{aero_blue}{RGB}{193, 234, 213}
\definecolor{pixie_green}{RGB}{183,214,170}
\definecolor{french_pass}{RGB}{195,232,246}
\definecolor{ice_cold}{RGB}{169,232,220}
\definecolor{pale_turquoise}{RGB}{172,240,242}
\definecolor{cruise}{RGB}{179,226,205}
\definecolor{sail}{RGB}{163,205,235}
\definecolor{spindle}{RGB}{179,205,227}
\definecolor{link_water}{RGB}{221, 232, 250}
\definecolor{periwinkle}{RGB}{203,213,232}
\definecolor{zanah}{RGB}{220, 233, 213}
\definecolor{frostee}{RGB}{217, 231, 214}
\definecolor{opal}{RGB}{199, 221, 211}
\definecolor{jet_stream}{RGB}{188, 214, 210}
\definecolor{skeptic}{RGB}{153, 187, 167}
\definecolor{hint_green}{RGB}{226,246,209}
\definecolor{snow_flurry}{RGB}{230,245,201}
\definecolor{surf_crest}{RGB}{205,230,208}
\definecolor{yellow_green}{RGB}{198,222,119}
\definecolor{cream}{RGB}{255,255,204}
\definecolor{pale_prim}{RGB}{255,255,179}
\definecolor{spring_sun}{RGB}{242,243,195}
\definecolor{portafino}{RGB}{245,237,160}
\definecolor{buttermilk}{RGB}{255,242,174}
\definecolor{cream_brulee}{RGB}{255, 229, 151}
\definecolor{dairy_cream}{RGB}{254,226,189}
\definecolor{champagne}{RGB}{254,217,166}
\definecolor{chardonnay}{RGB}{250,196,114}
\definecolor{manhattan}{RGB}{226,180,125}
\definecolor{rajah}{RGB}{253,180,98}
\definecolor{early_dawn}{RGB}{252,243,218}
\definecolor{egg_shell}{RGB}{238, 234, 215}
\definecolor{selago}{RGB}{243, 232, 243}
\definecolor{quartz}{RGB}{219,223,238}
\definecolor{fog}{RGB}{213, 193, 234}
\definecolor{languid_lavender}{RGB}{222,203,228}
\definecolor{watusi}{RGB}{254,221,207}
\definecolor{coral_andy}{RGB}{243,204,205}
\definecolor{cosmos}{RGB}{248,209,210}
\definecolor{melon}{RGB}{254,191,181}
\definecolor{azalea}{RGB}{234, 193, 194}
\definecolor{beauty_bush}{RGB}{235, 185, 179}
\definecolor{sundown}{RGB}{249, 180, 181}
\definecolor{mona_lisa}{RGB}{246,152,134}
\definecolor{salmon}{RGB}{242,131,107}

\definecolor{summer_sky}{RGB}{58, 151, 233}
\definecolor{chateau_green}{RGB}{72, 179, 96}
\definecolor{matisse}{RGB}{25, 104, 167}
\definecolor{allports}{RGB}{31, 106, 125}
\definecolor{sun_shade}{RGB}{255, 144, 68}
\definecolor{flamingo}{RGB}{237, 88, 85}
\definecolor{studio}{RGB}{128, 91, 160}

\definecolor{maya_blue}{RGB}{102, 204, 255}
\definecolor{feijoa}{RGB}{178,223,138}
\definecolor{sushi}{RGB}{117, 168, 47}
\definecolor{norway}{RGB}{158, 194, 132}
\definecolor{japanese_laurel}{RGB}{53, 116, 40}
\definecolor{see_green}{RGB}{161,228,195}
\definecolor{monte_carlo}{RGB}{135,204,194}
\definecolor{granny_smith_apple}{RGB}{150,214,150}
\definecolor{moss_green}{RGB}{170,216,176}
\definecolor{chateau_green}{RGB}{72, 179, 96}
\definecolor{opal}{RGB}{164,207,190}
\definecolor{acapulco}{RGB}{117, 170, 148}
\definecolor{viridian}{RGB}{55, 137, 122}
\definecolor{amazon}{RGB}{56, 123, 84}
\definecolor{asparagus}{RGB}{123, 160, 91}
\definecolor{fruit_salad}{RGB}{91, 160, 94}
\definecolor{puerto_rico}{RGB}{72, 179, 150}
\definecolor{mountain_meadow}{RGB}{0, 163, 136}
\definecolor{matisse}{RGB}{25, 104, 167}
\definecolor{allports}{RGB}{31, 106, 125}
\definecolor{astral}{RGB}{55, 111, 137}
\definecolor{spring_leaves}{RGB}{46, 83, 117}
\definecolor{biscay}{RGB}{44, 62, 80}
\definecolor{midnight}{RGB}{0, 29, 50}
\definecolor{amethyst}{RGB}{153, 102, 204}
\definecolor{studio}{RGB}{128, 91, 160}
\definecolor{tapestry}{RGB}{194, 109, 132}
\definecolor{atomic_tangerine}{RGB}{255, 153, 102}
\definecolor{amber}{RGB}{255, 191, 0}
\definecolor{casablanca}{RGB}{244, 178, 84}
\definecolor{california}{RGB}{233, 140, 58}
\definecolor{tomato}{RGB}{255, 97, 56} 
\definecolor{alizarin}{RGB}{233, 58, 64}

\definecolor{linen}{RGB}{251, 239, 227}
\definecolor{double_pearl_lusta}{RGB}{253, 242, 208}
\definecolor{oasis}{RGB}{253, 242, 208}
\definecolor{milan}{RGB}{255, 254, 169}
\definecolor{texas}{RGB}{245, 232, 123}
\definecolor{maize}{RGB}{249, 212, 156}

\definecolor{turmeric}{RGB}{211, 178, 76}
\definecolor{saffron}{RGB}{249,193,62}
\definecolor{my_sin}{RGB}{255, 176, 59}
\definecolor{tree_poppy}{RGB}{246, 154, 27}
\definecolor{jaffa}{RGB}{240, 131, 58}
\definecolor{crusta}{RGB}{254, 127, 44}
\definecolor{tahiti_gold}{RGB}{223, 102, 36}
\definecolor{outrageous_orange}{RGB}{255, 100, 45}
\definecolor{safety_orange}{RGB}{254, 106, 0}

\definecolor{azalea}{RGB}{251, 196, 196}
\definecolor{oyster_pink}{RGB}{238,206,205} 
\definecolor{coral_candy}{RGB}{242,208,205} 
\definecolor{baby_pink}{RGB}{246, 194, 192}
\definecolor{petite_orchid}{RGB}{223, 157, 155}
\definecolor{apricot}{RGB}{241,140,122}
\definecolor{NY_pink}{RGB}{228,136,113}
\definecolor{carmine_pink}{RGB}{231, 76, 60}
\definecolor{deep_carmine_pink}{RGB}{236, 50, 67}

\definecolor{wewak}{RGB}{244, 143, 150}
\definecolor{light_coral}{RGB}{244, 127, 123}
\definecolor{bittersweet}{RGB}{255,111,105}
\definecolor{carnation}{RGB}{245, 80, 86}
\definecolor{flamingo}{RGB}{237, 88, 85}
\definecolor{sunset_orange}{RGB}{242,89,75}
\definecolor{ku_crimson}{RGB}{243, 0, 25}
\definecolor{amaranth}{RGB}{234,46,73}
\definecolor{valencia}{RGB}{214, 87, 70}
\definecolor{chilean_fire}{RGB}{215, 87, 44}
\definecolor{mexican_red}{RGB}{170, 41, 37}

\definecolor{napa}{RGB}{163, 154, 137}

\definecolor{athens_gray}{RGB}{236, 240, 241}
\definecolor{gallery}{RGB}{240,240,240}
\definecolor{mercury}{RGB}{230,230,230}
\definecolor{platinum}{RGB}{228,228,228}
\definecolor{silver}{RGB}{191,191,191}
\definecolor{aluminum}{RGB}{153,153,153}
\definecolor{ship_gray}{RGB}{77,77,77}
\definecolor{tuatara}{RGB}{67, 67, 67}

\definecolor{malibu}{RGB}{110, 180, 240}
\definecolor{celestial_blue}{RGB}{52, 152, 219}
\definecolor{curious_blue}{RGB}{41, 128, 185}
\definecolor{french_blue}{RGB}{0, 112, 182}
\definecolor{matisse}{RGB}{25, 104, 167}
\definecolor{shakespeare}{RGB}{85, 154, 193}
\definecolor{seagull}{RGB}{128,177,211}
\definecolor{jelly_bean}{RGB}{45, 126, 150}
\definecolor{venice_blue}{RGB}{87, 135, 105}
\definecolor{boston_blue}{RGB}{68, 147, 161}

\definecolor{turquoise}{RGB}{41,217,194}
\definecolor{java}{RGB}{2,190,196}
\definecolor{riptide}{RGB}{141,211,199}
\definecolor{mountain_meadow}{RGB}{0, 163, 136}
\definecolor{free_speech_aquamarine}{RGB}{0, 156, 114}

\definecolor{cosmic_latte}{RGB}{222, 247, 229}
\definecolor{chinook}{RGB}{163, 232, 178}
\definecolor{padua}{RGB}{121, 189, 143}
\definecolor{ocean_green}{RGB}{79, 176, 112}
\definecolor{pastel_green}{RGB}{107, 227, 135}
\definecolor{chateau_green}{RGB}{69, 191, 85}
\definecolor{RoyalBlue}{RGB}{69, 191, 85}
\definecolor{pigment_green}{RGB}{0, 175, 79}
\definecolor{fern}{RGB}{101,197,117}
\definecolor{killarney}{RGB}{56, 113, 66}
\makeatletter
\tikzset{pics/named scope code/.style={code={\tikz@fig@mustbenamed%
			\begin{scope}[local bounding box/.expanded=\tikz@fig@name]#1\end{scope}%
}}}
\makeatother

\usetikzlibrary{positioning,shapes.misc,shapes.geometric}
\usetikzlibrary{decorations.pathreplacing}
\usetikzlibrary{arrows.meta}
\usetikzlibrary{fit}
\usetikzlibrary{backgrounds}

\definecolor{color-model}{RGB}{212,239,248}
\definecolor{color-layer}{RGB}{245,133,132}
\definecolor{color-h-square}{RGB}{246,192,115}
\definecolor{color-origin}{RGB}{135,204,194}
\definecolor{color-ice}{RGB}{185,205,246}

\usepackage{times}
\usepackage{latexsym}

\usepackage[T1]{fontenc}

\usepackage[utf8]{inputenc}

\usepackage{microtype}

\newcommand{\hide}[1] %

\title{IDPG: An Instance-Dependent Prompt Generation Method}

\author{Zhuofeng Wu$^{1}$\thanks{\hspace{1.5 mm} Work partially done while interning at Meta AI.}\hspace{6 mm} 
Sinong Wang$^2
$\hspace{6 mm} Jiatao Gu$^2$ \hspace{6 mm} Rui Hou$^{2}$\\ \textbf{Yuxiao Dong$^{2,3}$\thanks{\hspace{1.5 mm} Work done when at Meta AI.}\hspace{6 mm} V.G.Vinod Vydiswaran$^{1,4}$ \hspace{6 mm} Hao Ma$^{2}$}\\
$^1$School of Information, University of Michigan \\ 
$^2$Meta AI \hspace{6 mm} 
$^3$Tsinghua University \\
$^4$Department of Learning Health Sciences, University of Michigan \\
\tt{\{zhuofeng, vgvinodv\}@umich.edu} \\
\tt{\{sinongwang, jgu, rayhou, haom\}@fb.com}\\
\tt{yuxiaod@tsinghua.edu.cn}}

\begin{document}
\maketitle
\begin{abstract}
Prompt tuning is a new, efficient NLP transfer learning paradigm that adds a task-specific prompt in each input instance during the model training stage. It freezes the pre-trained language model and only optimizes a few task-specific prompts. In this paper, we propose a conditional prompt generation method to generate prompts for each input instance, referred to as the Instance-Dependent Prompt Generation (IDPG). Unlike traditional prompt tuning methods that use a fixed prompt, IDPG introduces a lightweight and trainable component to generate prompts based on each input sentence. Extensive experiments on ten natural language understanding (NLU) tasks show that the proposed strategy consistently outperforms various prompt tuning baselines and is on par with other efficient transfer learning methods such as Compacter while tuning far fewer model parameters.
\end{abstract}
\section{Introduction} \label{introduction}

Recently, pre-training a transformer model on a large corpus with language modeling tasks and fine-tuning it on different downstream tasks has become the main transfer learning paradigm in natural language processing~\citep{devlin2019bert}. 
Notably, this paradigm requires updating and storing all the model parameters for every downstream task. 
As the model size proliferates (e.g., 330M parameters for BERT~\citep{devlin2019bert} and 175B for GPT-3~\citep{brown2020language}), it becomes computationally expensive and challenging to fine-tune the entire pre-trained language model (LM). 
Thus, it is natural to ask the question of whether we can transfer the knowledge of a pre-trained LM into downstream tasks by  tuning only a small portion of its parameters with most of them freezing. 

Studies have attempted to address this question from different perspectives. 
One line of research~\citep{li2021prefix} suggests to augment the model with a few small trainable modules and freeze the original transformer weight. 
Take Adapter~\citep{houlsby2019parameter, pfeiffer2020adapterfusion, pfeiffer2020mad} and Compacter~\citep{mahabadi2021compacter} for example, both of them insert a small set of additional modules between each transformer layer. 
During fine-tuning, only these additional and task-specific modules are trained, reducing the trainable parameters to $\sim 1$--3\% of the original transformer model per task.

Another line of works focus on prompting. 
The GPT-3 models~\citep{brown2020language, schick2020exploiting} find that with proper manual prompts, a 
pre-trained LM can successfully match the fine-tuning performance of BERT models. 
LM-BFF~\citep{gao2020making}, EFL~\citep{wang2021entailment}, and AutoPrompt~\citep{shin2020autoprompt} further this direction by insert prompts in the input embedding layer. 
However, these methods rely on grid-search for a natural language-based prompt from a large search space, resulting in difficulties to optimize. 

To tackle this issue, prompt tuning~\citep{lester2021power}, prefix tuning~\citep{li2021prefix}, and P-tuning~\citep{liu2021p, liu2021gpt} are proposed to prepend trainable prefix tokens to the input layer and train these soft prompts only during the fine-tuning stage. 
In doing so, the problem of searching discrete prompts are converted into an continuous optimization task, which can be solved by a variety of optimization techniques such as SGD  %
and thus significantly reduced the number of trainable parameters to only a few thousand. 
However, all existing prompt-tuning methods have thus far focused on  task-specific prompts, making them incompatible with the traditional LM objective. 
For example, it is unlikely to see many different sentences with the same prefix in the pre-training corpus. Thus, a unified prompt may disturb the prediction and lead to a performance drop. 
In light of these limitations, we instead ask the following question:~\emph{Can we generate input-dependent prompts to smooth the domain difference?}

In this paper, we present the instance-dependent prompt generation (IDPG) strategy for efficiently tuning large-scale LMs. 
Different from the traditional prompt-tuning methods that rely on a fixed prompt for each task, IDPG instead develops a conditional prompt generation model to generate prompts for each instance. 
Formally, the IDPG generator can be denoted as $f(x;\mathbf{W})$, where $x$ is the instance representation  and $\mathbf{W}$ represents the trainable parameters. 
Note that by setting $\mathbf{W}$ to a zero matrix and only training the bias, IDPG would degenerate into the traditional prompt tuning process~\cite{lester2021power}. 
To further reduce the number of parameters in the generator $f(x;\mathbf{W})$, we propose to apply a lightweight bottleneck architecture (i.e., a two-layer perceptron) and then decompose it by a parameterized hypercomplex multiplication (PHM) layer~\citep{zhang2021beyond}. 
To summarize, this works makes the following contributions:  %
\begin{itemize}
    \item We introduce an input-dependent prompt generation method---IDPG---that  only requires training 134K parameters per task, corresponding to $\sim$0.04\% of a pre-trained LM such as RoBERTa-Large~\citep{liu2019roberta}.
    \item Extensive evaluations on ten natural language understanding (NLU) tasks show that IDPG consistently outperforms task-specific prompt tuning methods by 1.6--3.1 points (Cf. Table~\ref{tab:main}). 
    Additionally, it also offers comparable performance to Adapter-based methods while using much fewer parameters (134K vs.~1.55M).
    \item We conduct substantial intrinsic studies, revealing how and why each component of the proposed model and the generated prompts could help the downstream tasks.  
\end{itemize}

\hide{ %

\section{Introduction} \label{introduction}

Pre-training a transformer model on a large corpus with language modeling tasks and fine-tuning it on different downstream tasks has become the main transfer learning paradigm in natural language processing over the past few years~\citep{devlin2019bert}. However, one notable problem for such a paradigm is that it requires updating and storing all the model parameters for every downstream task. As the model size proliferates (e.g., BERT~\citep{devlin2019bert} has 330M parameters while GPT-3~\citep{brown2020language} has 175B parameters), it has become more costly and challenging to fine-tune the entire pre-trained language model (LM). One natural question is that can we transfer the knowledge of pre-trained LM into downstream tasks by freezing most of the model parameters and only tuning a small portion of them.

\begin{figure}
\centering
\resizebox{0.98\linewidth}{!}{
    \begin{tikzpicture}[]
	\begin{axis}[
	   nodes near coords, 
	   every node near coord/.append style={xshift=+1.0mm,anchor=west},
	   ymin=88.5, ymax=92.6,
	   xmode=log,
	   enlargelimits=0.2,
	   log ticks with fixed point,
	   xtick={0.00001, 0.001, 0.01, 0.1, 1, 10, 100},
	   tick label style={font=\scriptsize},
	   xlabel=\small Percentage of trained parameters per task (relative to RoBERTa),
       ylabel=\small Average Score,
       ymajorgrids=true,
       xmajorgrids=true,
       grid=major, %
       major grid style={gallery!50},
       axis background/.style={fill=white!10},
       ]

        \addplot[
        mark=*,
        only marks,
        draw opacity=0.5,
        fill=maya_blue,
        mark size=4pt,
        point meta=explicit symbolic
        ]
        coordinates {
        (0.00137544, 88.8) [prompt tuning]
        };
        
        \addplot[
        mark=*,
        only marks,
        draw opacity=0.5,
        fill=maya_blue,
        mark size=4pt,
        point meta=explicit symbolic
        ]
        coordinates {
        (0.03301, 90.3) [P-tuning v2]
        };

        \addplot[
        mark=*,
        only marks,
        draw opacity=0.5,
        fill=chateau_green,
        mark size=4pt,
        point meta=explicit symbolic,
        ]
        coordinates {
        (0.043463, 92.2) [Compacter]
        };
        
        \addplot[
        mark=*,
        only marks,
        draw opacity=0.5,
        fill=flamingo,
        mark size=4pt,
        point meta=explicit symbolic
        ]
        coordinates {
        (0.4264, 92.6) [Adapter]
        };
        
        \addplot[
        mark=*,
        draw opacity=0.5,
        fill=summer_sky,
        only marks,
        mark size=4pt,
        point meta=explicit symbolic,
        ]
        coordinates {
        (0.036862, 91.9) [IDPG-PHM]
        };
        
        \addplot[
        mark=*,
        draw opacity=0.5,
        fill=summer_sky,
        only marks,
        mark size=4pt,
        point meta=explicit symbolic,
        ]
        coordinates {
        (0.038787, 91.6) [IDPG-PHM-GloVe]
        };
        
        \addplot[
        mark=*,
        draw opacity=0.5,
        only marks,
        fill=melon,
        point meta=explicit symbolic,
        mark size=4pt,
        every node near coord/.append style={yshift=-0.4mm,anchor=north, font=\small}
        ]
        coordinates {
        (100, 92.2) [RoBERTa-FT]
        };
        \draw[dotted, thick, draw=flamingo] (-8.8, 385) rectangle (-1.1, 460);

        \end{axis}
\end{tikzpicture}}
    \caption{Overall evaluation on ten NLU tasks, wherein parameters from classification heads are not included.}
    \vspace{-0.1in}
    \label{fig:overall-performance}
\end{figure}

There are several lines of research investigating efficient methods to fine-tune a pre-trained LM. One line of research~\citep{li2021prefix} shows that we can augment the model with a few small trainable modules and freeze the original transformer weight. For example, Adapter~\citep{houlsby2019parameter, pfeiffer2020adapterfusion, pfeiffer2020mad} and Compacter~\citep{mahabadi2021compacter} insert a small set of additional modules between each transformer layer. They only train these additional, task-specific modules during fine-tuning, which efficiently reduces the trainable parameters to 1--3\% of original transformer model per task.

Another line of research focuses on prompting. GPT-3 models~\citep{brown2020language, schick2020exploiting} first suggested that with proper manual prompt, large-scale pre-trained LM can successfully match the fine-tuning performance of BERT models. LM-BFF~\citep{gao2020making}, EFL~\citep{wang2021entailment}, and AutoPrompt~\citep{shin2020autoprompt} started transforming the standard fine-tuning approach into a new training paradigm with a prompt inserted in the input embedding layer. These works focus on grid-search of a natural language-based prompt, which creates a large search space and is usually difficult to optimize. 
Instead, prompt tuning~\citep{lester2021power}, prefix tuning~\citep{li2021prefix}, and P-tuning~\citep{liu2021gpt} models were proposed to prepend trainable prefix tokens to the input layer and only train these soft prompts during the fine-tuning stage. This framed the problem of searching discrete prompts into an continuous optimization task, and allowed use of a variety of optimization techniques such as SGD to find the optimal prompts. These techniques significantly reduced the number of trainable parameters to only a few thousand. However, all existing prompt-tuning methods focus on a task-specific prompt and are actually incompatible with the traditional LM objective. For example, it is unlikely to see many different sentences with the same prefix in the pre-training corpus. This leads to the following question:~\emph{Can we generate an input-dependent prompt to smooth the domain difference?}

In this paper, we propose a novel approach we refer to as IDPG -- instance-dependent prompt generation. Unlike the traditional prompt-tuning which relies on a fixed prompt for each task, IDPG leverages a conditional prompt generation model to generate a prompt for each instance. One can regard our generator as $f(x;\mathbf{W})$ where $x$ is the instance representation, and $\mathbf{W}$ represents the trainable parameters. In the extreme case, setting $\mathbf{W}$ to a zero matrix and only training the bias would degenerate it to the traditional prompt tuning~\cite{lester2021power}. To further reduce the number of parameters in the generator $f(x;\mathbf{W})$, we first apply a lightweight bottleneck architecture (a two-layer perceptron), and then decompose it by a parameterized hypercomplex multiplication (PHM) layer~\citep{zhang2021beyond}. In summary, our contributions are three-fold:
\begin{itemize}
    \item We proposed an input-dependent prompt generation method, IDPG, which only requires training 134K parameters per task (roughly equal to 0.037\% of a pre-trained LM like RoBERTa-Large~\citep{liu2019roberta}).
    \item A systematic evaluation on ten natural language understanding (NLU) tasks shows that our proposed method consistently outperforms the traditional task-specific prompt tuning methods by 1.6--5.0 points. Our method also has comparable performance to Adapter-based methods while using much fewer parameters (134K vs.~6.2M).
    \item We provided substantial ablation studies, revealing the benefits of instance-dependent prompts and the impact of each component. 
\end{itemize}

}%

\section{Preliminary} \label{preliminary}

\subsection{Manual Prompt} \label{efl}

Manual prompt learning~\cite{brown2020language,schick2020exploiting} inserts a pre-defined label words in each input sentence. For example, it reformulates a sentence sentiment classification task with an input sentence $S_1$ as 
\[
x_{in} = \texttt{[CLS]} P \texttt{[SEP]} S_1 \texttt{[EOS]},
\]
where $P$ is the prompt such as ``indicating the positive user sentiment''. Using the pre-trained language model $\mathbf{M}$, we can obtain the sentence representation $\mathbf{h}_{\texttt{[CLS]}}=\mathbf{M}(x_{in})$, and train a task-specific head $\texttt{softmax}(\mathbf{W}\mathbf{h}_{\texttt{[CLS]}})$ to maximize the log-probability of the correct label. LM-BFF \citep{gao2020making} shows that adding a specifically designed prompt during fine-tuning can benefit the few-shot scenario. EFL~\citep{wang2021entailment} further suggests that reformulating the task as entailment can further improve the performance in both low-resource and high-resource scenarios.

\subsection{Prompt Tuning} \label{prompt}

Prompt tuning~\cite{lester2021power}, prefix tuning~\cite{li2021prefix}, and P-tuning~\cite{liu2021p, liu2021gpt} methods propose to insert a trainable prefix in front of the input sequence. Specifically, they reformulate the input for single sentence tasks as
\begin{align*}
x_{in} =\texttt{concat}[\mathbf{W}_p, \mathbf{E}(\texttt{[SEP]}S_2 \texttt{[EOS]})]
\end{align*}
and for sentence pair tasks as
\begin{align*}
x_{in} = \texttt{concat}[\mathbf{W}_p, \mathbf{E}(\texttt{[SEP]}S_2\texttt{[SEP]} S_3\texttt{[EOS]})],
\end{align*}
where $\mathbf{W}_p$ is the embedding table of the inserted prompt, $S_2$ and $S_3$ are input sentences, and $\mathbf{E}$ denotes the operation of tokenization and extraction of embeddings. Apart from LM-BFF and EFL, there is no corresponding real text for the prompt as $\mathbf{W}_p$ is a set of random-initialized tensors to represent the soft prompt.

\tikzset{
		larrow/.style={arrows={-{Latex[length=1.25mm, width=1.mm]}}, thick},
		lrarrow/.style={arrows={{Latex[length=1.25mm, width=1.mm]}-{Latex[length=1.25mm, width=1.mm]}}, thick},
		square/.style={
			rectangle,inner sep=-0.5mm, outer sep=0,minimum width=1.8em,minimum height=1.6em,align=center,font=\tiny,fill=color-ice,rounded corners=1},
		origin/.style={
			regular polygon, regular polygon sides=6,minimum width=1.9em, align=center, inner sep=-0.5mm, outer sep=0,font=\tiny,fill=color-origin,rounded corners=1},
		model/.style={draw=black,fill=color-model,fill opacity=1,inner sep=0.5em, rounded corners=3,label={center:\scriptsize Pre-trained Model}},
		model1/.style={densely dotted, draw=black,fill=color-model,fill opacity=1,inner sep=0.5em, rounded corners=3},
		model2/.style={rectangle,draw,inner xsep=1pt,outer sep=0pt,minimum height=1mm,fill=color-layer,fill opacity=1,rounded corners=0.5,text width=1.5cm,execute at begin node=\setlength{\baselineskip}{2pt},anchor=mid,align=center,},
		model3/.style={dashed,draw=black,fill=color-model,fill opacity=1,inner sep=0.4em, rounded corners=3,minimum width=3.5cm,},
		layer/.style={rectangle,align=center,inner xsep=2pt,minimum height=8mm,fill=color-layer,font=\scriptsize,rounded corners=1},
		layer1/.style={draw=black,fill=color-layer,fill opacity=1,inner sep=0.5em, label={center:\scriptsize Prompt Generator}},
		pics/.cd,
		circle0/.style={named scope code={
				\pgfmathsetmacro{\r}{0.135}
				\filldraw[fill=color-ice,fill opacity=1,rounded corners=0.5] (-8.25*\r,-1.5*\r) rectangle (8.25*\r,1.5*\r);
				\foreach \x in {-6.25,-3.75,...,6.25}
				\filldraw[line width=0.3pt,fill=linen] (\x*\r,0) circle (\r);
		}},
		circle1/.style={named scope code={
			\pgfmathsetmacro{\r}{0.135}
			\filldraw[fill=color-layer,fill opacity=1,rounded corners=0.5] (-8.25*\r,-1.5*\r) rectangle (8.25*\r,1.5*\r);
			\foreach \x in {-6.25,-3.75,...,6.25}
			\filldraw[line width=0.3pt,fill=white] (\x*\r,0) circle (\r);
		}},
		circle2/.style={named scope code={
			\pgfmathsetmacro{\r}{0.135}
			\filldraw[fill=color-layer,fill opacity=1,rounded corners=0.5] (-4.5*\r,-1.5*\r) rectangle (4.5*\r,1.5*\r);
			\foreach \x in {-2.5,0,2.5}
			\filldraw[line width=0.3pt,fill=white] (\x*\r,0) circle (\r);
		}},
	}
\begin{figure*}[!ht]
		\centering
		\captionsetup[subfigure]{width=0.45\textwidth}
		\subcaptionbox{Manual Prompt}[.5\linewidth]{
			\resizebox{0.96\linewidth}{!}{
				\begin{tikzpicture}[scale=1, line width=0.6pt,node distance=0.5mm]
					\node[square] (e-CLS) at (0,0){$\bm{E}_{\mathrm{[CLS]}}$};
					\node[square,right=of e-CLS] (e-I) {$\bm{E}_{\mathrm{I}}$};
					\node[square,right=of e-I] (e-love) {$\bm{E}_{\mathrm{love}}$};
					\node[square,right=of e-love] (e-these) {$\bm{E}_{\mathrm{these}}$};
					\node[square,right=of e-these] (e-actors) {$\bm{E}_{\mathrm{actors}}$};
					\node[square,right=of e-actors] (e-SEP) {$\bm{E}_{\mathrm{[SEP]}}$};
					\node[square,right=of e-SEP] (e-they) {$\bm{E}_{\mathrm{they}}$};
					\node[square,right=of e-they] (e-are) {$\bm{E}_{\mathrm{are}}$};
					\node[square,right=of e-are] (e-great) {$\bm{E}_{\mathrm{great}}$};
					\node[square,right=of e-great] (e-EOS) {$\bm{E}_{\mathrm{[EOS]}}$};
					
					\node[origin,below=5mm of e-CLS] (CLS) {[CLS]};
					\node[origin,below=5mm of e-I] (I) {I};
					\node[origin,below=5mm of e-love] (love) {love};
					\node[origin,below=5mm of e-these] (these) {these};
					\node[origin,below=5mm of e-actors] (actors) {actors};
					\node[origin,below=5mm of e-SEP] (SEP) {[SEP]};
					\node[origin,below=5mm of e-they] (they) {they};
					\node[origin,below=5mm of e-are] (are) {are};
					\node[origin,below=5mm of e-great] (great) {great};
					\node[origin,below=5mm of e-EOS] (EOS) {[EOS]};
					
					\draw[larrow] (CLS) -- (e-CLS);
					\draw[larrow] (I) -- (e-I);
					\draw[larrow] (love) -- (e-love);
					\draw[larrow] (these) -- (e-these);
					\draw[larrow] (actors) -- (e-actors);
					\draw[larrow] (SEP) -- (e-SEP);
					\draw[larrow] (they) -- (e-they);
					\draw[larrow] (are) -- (e-are);
					\draw[larrow] (great) -- (e-great);
					\draw[larrow] (EOS) -- (e-EOS);
					
					\node [densely dotted, draw=black, fit={(they) (great)}, inner sep=0.4] (dotted) {};
					\draw[thick,draw=black!42,decorate,decoration={brace,mirror,raise=1pt}] (dotted.south west) --node[shift={(-90:10pt)}]{\scriptsize manually generated} node[shift={(-90:18pt)}]{\scriptsize prompt} (dotted.south east);
					
					\node[square,above=0.5cm of e-CLS,fill=color-h-square] (h-CLS) {$\bm{h}_{\mathrm{[CLS]}}$};
					\node[square,above=0.5cm of e-I,] (h-2) {};
					\node[square,above=0.5cm of e-love,] (h-3) {};
					\node[square,above=0.5cm of e-these,] (h-4) {};
					\node[square,above=0.5cm of e-actors,] (h-5) {};
					\node[square,above=0.5cm of e-SEP,] (h-6) {};
					\node[square,above=0.5cm of e-they,] (h-7) {};
					\node[square,above=0.5cm of e-are,] (h-8) {};
					\node[square,above=0.5cm of e-great,] (h-9) {};
					\node[square,above=0.5cm of e-EOS,] (h-10) {};
					
					\begin{scope}[on background layer]
						\node[model,fit={(h-CLS) (e-EOS)}] (pre) {};
					\end{scope}
					
					\node[layer,outer sep=0,above=17mm of pre.west,anchor=west] (ent) {Classification head \\ \scriptsize Feed forward layer};
					\draw[larrow] (h-CLS.north) -- (h-CLS.north |- ent.south);
					
					\node[draw,rectangle,align=left,inner xsep=3pt,outer sep=0,minimum width=15mm,minimum height=5mm,font=\scriptsize,right=30mm of ent,rounded corners=1] (surt) {\hspace{-6mm}\,$\surd$ \;\scriptsize 1 \\ {\large \hspace{-6mm}$\times$} \scriptsize 0};
					
					\draw[larrow] (ent) --node[shift={(90:5pt)}]{\scriptsize Predict} (surt);
			\end{tikzpicture}}
		}%
		\subcaptionbox{Prompt Tuning}[.45\linewidth]{
			\resizebox{0.96\linewidth}{!}{
				\begin{tikzpicture}[scale=1, line width=0.6pt, node distance=0.5mm]
					\node[square] (e-CLS) at (0,0){$\bm{E}_{\mathrm{[CLS]}}$};
					\node[square,fill=color-layer,right=of e-CLS] (e-1) {$\bm{E}_{1}$};
					\node[square,fill=color-model,right=-1.5mm of e-1] (e-2) {$\cdots$};
					\node[square,fill=color-layer,right=-1.5mm of e-2] (e-3) {$\bm{E}_{t}$};
					\node[square,fill=color-layer,right=of e-CLS] (e-1) {$\bm{E}_{1}$};
					\node[square,right=of e-3] (e-I) {$\bm{E}_{\mathrm{I}}$};
					\node[square,right=of e-I] (e-love) {$\bm{E}_{\mathrm{love}}$};
					\node[square,right=of e-love] (e-these) {$\bm{E}_{\mathrm{these}}$};
					\node[square,right=of e-these] (e-actors) {$\bm{E}_{\mathrm{actors}}$};
					\node[square,right=of e-actors] (e-EOS) {$\bm{E}_{\mathrm{[EOS]}}$};
					
					\node[origin,below=3.0mm of e-CLS] (CLS) {[CLS]};
					\node[origin,below=3.0mm of e-I] (I) {I};
					\node[origin,below=3.0mm of e-love] (love) {love};
					\node[origin,below=3.0mm of e-these] (these) {these};
					\node[origin,below=3.0mm of e-actors] (actors) {actors};
					\node[origin,below=3.0mm of e-EOS] (EOS) {[EOS]};
					
					\draw[larrow] (CLS) -- (e-CLS);
					\draw[larrow] (I) -- (e-I);
					\draw[larrow] (love) -- (e-love);
					\draw[larrow] (these) -- (e-these);
					\draw[larrow] (actors) -- (e-actors);
					\draw[larrow] (EOS) -- (e-EOS);
					
					\node[square,above=0.41cm of e-CLS,fill=color-h-square] (h-CLS) {$\bm{h}_{\mathrm{[CLS]}}$};
					\node[square,above=0.41cm of e-1,] (h-2) {};
					\node[right=-0.05cm of h-2] (h-2) {\tiny $\cdots$};
					\node[square,above=0.41cm of e-3,] (h-3) {};
					\node[square,above=0.41cm of e-I,] (h-4) {};
					\node[square,above=0.41cm of e-love,] (h-5) {};
					\node[square,above=0.41cm of e-these,] (h-6) {};
					\node[square,above=0.41cm of e-actors,] (h-7) {};
					\node[square,above=0.41cm of e-EOS,] (h-8) {};
					\node[layer,outer sep=0,above=17mm of pre.west,anchor=west] (cla) {Classification head \\ \scriptsize Feed forward layer};
					\draw[larrow] (h-CLS.north) -- (h-CLS.north |- cla.south);
					
					\node[draw,rectangle,align=left,inner xsep=3pt,outer sep=0,minimum width=15mm,minimum height=5mm,font=\scriptsize,right=30mm of ent,rounded corners=1] (surt) {\hspace{-6mm}\,$\surd$ \;\scriptsize 1 \\ {\large \hspace{-6mm}$\times$} \scriptsize 0};
					
					\draw[larrow] (cla) --node[shift={(90:5pt)}]{\scriptsize Predict} (surt);
					
					\begin{scope}[on background layer]
						\node[model,fit={(h-CLS) (e-EOS)}] (pre) {};
						\node[fill=color-ice, fit={(e-1) (e-3)},rounded corners=1,inner sep=0] (e1-3) {};
					\end{scope}
					
					\draw[thick,draw=black!42,decorate,decoration={brace,mirror,raise=1pt}] ([yshift=-10mm]e1-3.south west) --node[shift={(-90:10pt)}]{\scriptsize randomly generated} node[shift={(-90:18pt)}]{\scriptsize prompt} ([yshift=-10mm]e1-3.south east);
					
			\end{tikzpicture}}
		}
		
		\subcaptionbox{Instance-Dependent Prompt Generation}[1\linewidth]{
			\resizebox{0.85\linewidth}{!}{
				\begin{tikzpicture}[scale=1, line width=0.6pt,
					node distance=0.5mm]
					\node[square] (e-CLS-2) at (0,0){$\bm{E}_{\mathrm{[CLS]}}$};
					\node[square,right=of e-CLS-2] (e-I-2) {$\bm{E}_{\mathrm{I}}$};
					\node[square,right=of e-I-2] (e-love-2) {$\bm{E}_{\mathrm{love}}$};
					\node[square,right=of e-love-2] (e-these-2) {$\bm{E}_{\mathrm{these}}$};
					\node[square,right=of e-these-2] (e-actors-2) {$\bm{E}_{\mathrm{actors}}$};
					\node[square,right=of e-actors-2] (e-SEP-2) {$\bm{E}_{\mathrm{[SEP]}}$};
					\node[square,fill=color-layer,right=of e-SEP-2] (e-1-2) {};
					\node[square,fill=color-model,right=-1.5mm of e-1-2] (e-2-2) {$\cdots$};
					\node[square,fill=color-layer,right=-1.5mm of e-2-2] (e-3-2) {$\bm{E}_{\mathrm{1,t}}$};
					\node[square,fill=color-layer,right=of e-SEP-2] (e-1-2) {$\bm{E}_{1,1}$};
					\node[square,right=of e-3-2] (e-EOS-2) {$\bm{E}_{\mathrm{[EOS]}}$};
					
					\node[origin,below=5mm of e-CLS-2] (CLS-2) {[CLS]};
					\node[origin,below=5mm of e-I-2] (I-2) {I};
					\node[origin,below=5mm of e-love-2] (love-2) {love};
					\node[origin,below=5mm of e-these-2] (these-2) {these};
					\node[origin,below=5mm of e-actors-2] (actors-2) {actors};
					\node[origin,below=5mm of e-SEP-2] (SEP-2) {[SEP]};
					\node[origin,below=5mm of e-EOS-2] (EOS-2) {[EOS]};
					
					\draw[larrow] (CLS-2) -- (e-CLS-2);
					\draw[larrow] (I-2) -- (e-I-2);
					\draw[larrow] (love-2) -- (e-love-2);
					\draw[larrow] (these-2) -- (e-these-2);
					\draw[larrow] (actors-2) -- (e-actors-2);
					\draw[larrow] (SEP-2) -- (e-SEP-2);
					\draw[larrow] (EOS-2) -- (e-EOS-2);
					
					\node[square,above=0.75cm of e-CLS-2,fill=color-h-square] (h-CLS-2) {$\bm{E}_{\mathrm{[CLS]}}$};
					\node[square,above=0.75cm of e-I-2] (e-I-3) {};
					\node[square,above=0.75cm of e-love-2] (e-love-3) {};
					\node[square,above=0.75cm of e-these-2] (e-these-3) {};
					\node[square,above=0.75cm of e-actors-2] (e-actors-3) {};
					\node[square,above=0.75cm of e-SEP-2] (e-SEP-3) {};
					\node[square,above=0.75cm of e-1-2,fill=color-layer] (e-1-3) {};
					\node[above=0.2cm of e-1-2] {\tiny $\cdots$};
					\node[above=0.2cm of e-2-2] {\tiny $\cdots$};
					\node[above=0.2cm of e-3-2] {\tiny $\cdots$};
					\node[square,fill=color-model,right=-1.5mm of e-1-3] (e-2-3) {$\cdots$};
					\node[square,above=0.75cm of e-1-2,fill=color-layer] (e-1-3) {$\bm{E}_{\mathrm{N,1}}$};
					\node[square,fill=color-layer,above=0.75cm of e-3-2] (e-3-3) {$\bm{E}_{\mathrm{N,t}}$};
					\node[square,above=0.75cm of e-EOS-2] (e-EOS-3) {};
					
					\begin{scope}[on background layer]
						\node[model,fit={(h-CLS-2) (e-EOS-2)}] (pre-2) {};
					\end{scope}
					
					\node[layer,outer sep=0,above=18mm of pre-2.west,anchor=west] (ent-2) {Classification head \\ \scriptsize Feed forward layer};
					\draw[larrow] (h-CLS-2.north) -- (h-CLS-2.north |- ent-2.south);
					
					\node[draw,rectangle,align=left,inner xsep=3pt,outer sep=0,minimum width=15mm,minimum height=5mm,font=\scriptsize,right=25mm of ent-2,rounded corners=1] (surt-2) {\hspace{-6mm}\,$\surd$ \;\scriptsize 1 \\ {\large \hspace{-6mm}$\times$} \scriptsize 0};
					\draw[larrow] (ent-2) --node[shift={(90:5pt)}]{\scriptsize Predict} (surt-2);
					
					\draw[thick,draw=black!42,decorate,decoration={brace,mirror,raise=1pt}] ([yshift=-11mm]e-1-2.south west) --node[shift={(-90:10pt)}]{\scriptsize Instance-dependent} node[shift={(-90:18pt)}]{\scriptsize prompt} ([yshift=-11mm]e-3-2.south east);
					
					\draw pic (c1) at(10.0,-3.56) {circle0};
					\node[below=0cm of c1] (E-CLS) {\scriptsize $\bm{M}{(x_i}$)};
					\node[model2,above=0.3cm of c1,] (w1) {\scriptsize down-project};
					\pic[above=0.4cm of w1] (c2) {circle2};
					\node[left=0cm of w1] {\scriptsize $w_{1}$};
					\node[model2,above=0.2cm of c2,] (N) {\scriptsize Nonlinearity};
					\node[model2,above=0.2cm of N,] (w2) {\scriptsize up-project};
					\node[left=0cm of w2] {\scriptsize $w_{2}$};
					\node[right=-0.1cm of w2,align=center,execute at begin node=\setlength{\baselineskip}{2pt},] {\tiny Prompt \\ \tiny Generator};
					
					\begin{scope}[on background layer]
						\node[model3,fit={(w1) (w2)}] (p) {};
					\end{scope}
					
					\node[above=0.5cm of w2] (cd1) {\scriptsize $\cdots$};
					\pic[left=1.1cm of cd1] (c3) {circle1};
					\node[left=-0.1cm of c3] {\scriptsize $\bm{E}_{1,1}$};
					\pic[right=1.1cm of cd1] (c4) {circle1};
					\node[right=-0.1cm of c4] {\scriptsize $\bm{E}_{\mathrm{1,t}}$};
					\node[above=0.27cm of cd1] (cd2) {\scriptsize $\cdots$};
					\node[left=0.8cm of cd2] (cd3) {\scriptsize $\cdots$};
					\node[right=0.8cm of cd2] (cd4) {\scriptsize $\cdots$};
					\node[above=0.92cm of cd1] (cd5) {\scriptsize $\cdots$};
					\pic[left=1.1cm of cd5] (c5) {circle1};
					\node[left=-0.1cm of c5] {\scriptsize $\bm{E}_{\mathrm{N,1}}$};
					\pic[right=1.1cm of cd5] (c6) {circle1};
					\node[right=-0.1cm of c6] {\scriptsize $\bm{E}_{\mathrm{N,t}}$};
					
					\draw[larrow] (c1.north) -- (w1.south);
					\draw[larrow] (w1.north) -- (c2.south);
					\draw[larrow] (c2.north) -- (N.south);
					\draw[larrow] (N.north) -- (w2.south);
					\draw[larrow] (w2.north) -- ++(0,0.3) -| (c3.south);
					\draw[larrow] (w2.north) -- ++(0,0.3) -| (c4.south);
					
					\draw[larrow,] (c3.north) -- ++(0,0.25)  -| (e-1-2.north);
					\draw[larrow,] (c4.north) -- ++(0,0.35)  -| (e-3-2.north);
					\draw[larrow,] (c5.north) -- ++(0,0.35)  -| (e-1-3.north);
					\draw[larrow,] (c6.north) -- ++(0,0.45)  -| (e-3-3.north);
					
					\fill[fill=color-layer,rounded corners=1]  (4.5,-3.5) rectangle node[shift={(0:17pt)}]{\scriptsize tuned} (5,-3);
					\fill[fill=color-ice,rounded corners=1]  (6,-3.5) rectangle node[shift={(0:18pt)}]{\scriptsize frozen} (6.5,-3);
			\end{tikzpicture}}
		}
		\vspace{-0.1in}
		\caption{An illustration of (a) manual prompt; (b) prompt-tuning method; (c) our proposed method. The red block refers to the trainable module, while the blue block refers to the frozen module.}
		\vspace{-0.2in}
		\label{fig:model}
	\end{figure*}

\hide{

\section{Preliminary} \label{preliminary}

\subsection{Manual Prompt} \label{efl}

Manual prompt learning~\cite{brown2020language,schick2020exploiting} insert a pre-defined label words in each input sentence. For example, it reformulates a sentence sentiment classification task with an input sentence $S_1$ as 
\[
x_{in} = \texttt{[CLS]} P \texttt{[SEP]} S_1 \texttt{[EOS]},
\]
where $P$ is the prompt such as ``indicating the positive user sentiment''. Using pre-trained language model $\mathbf{M}$, we can obtain the sentence representation $\mathbf{h}_{\texttt{[CLS]}}=\mathbf{M}(x_{in})$, and train a task-specific head $\texttt{softmax}(\mathbf{W}\mathbf{h}_{\texttt{[CLS]}})$ to maximize the log-probability of the correct label. LM-BFF \citep{gao2020making} showed that adding a specifically designed prompt during fine-tuning can benefit the few-shot scenario. EFL~\citep{wang2021entailment} further showed that reformulating the task as entailment can further improve the performance in both low-resource and high-resource scenarios.

\subsection{Prompt Tuning} \label{prompt}

Prompt tuning~\cite{lester2021power}, prefix tuning~\cite{li2021prefix}, and P-tuning~\cite{liu2021gpt} methods propose to insert a trainable prefix in front of the input sequence. Specifically, they reformulate the input for single sentence tasks as
\begin{align*}
x_{in} =\texttt{concat}[\mathbf{W}_p, \mathbf{E}(\texttt{[SEP]}S_2 \texttt{[EOS]})]
\end{align*}
and for sentence pair tasks as
\begin{align*}
x_{in} = \texttt{concat}[\mathbf{W}_p, \mathbf{E}(\texttt{[SEP]}S_2\texttt{[SEP]} S_3\texttt{[EOS]})],
\end{align*}
where $\mathbf{W}_p$ is the embedding table of the inserted prompt, $S_2$ and $S_3$ are input sentences, and $\mathbf{E}$ denotes the operation of tokenization and extraction of embedding. Apart from LM-BFF and EFL, there is no corresponding real text for the prompt as $\mathbf{W}_p$ is a set of random-initialized tensors to represent the soft prompt.

\tikzset{
	larrow/.style={arrows={-{Latex[length=1.25mm, width=1.mm]}}, thick},
	lrarrow/.style={arrows={{Latex[length=1.25mm, width=1.mm]}-{Latex[length=1.25mm, width=1.mm]}}, thick},
	square/.style={
		rectangle,inner sep=-0.5mm, outer sep=0,minimum width=1.8em,minimum height=1.6em,align=center,font=\tiny,fill=color-ice,rounded corners=1},
	origin/.style={
		regular polygon, regular polygon sides=6,minimum width=1.9em, align=center, inner sep=-0.5mm, outer sep=0,font=\tiny,fill=color-origin,rounded corners=1},
	model/.style={draw=black,fill=color-model,fill opacity=1,inner sep=0.5em, rounded corners=3,label={center:\scriptsize Pre-trained Model}},
	model1/.style={densely dotted, draw=black,fill=color-model,fill opacity=1,inner sep=0.5em, rounded corners=3},
	layer/.style={rectangle,align=center,inner xsep=2pt,minimum height=8mm,fill=color-layer,font=\scriptsize,rounded corners=1},
	layer1/.style={draw=black,fill=color-layer,fill opacity=1,inner sep=0.5em, label={center:\scriptsize Prompt Generator}},
}

	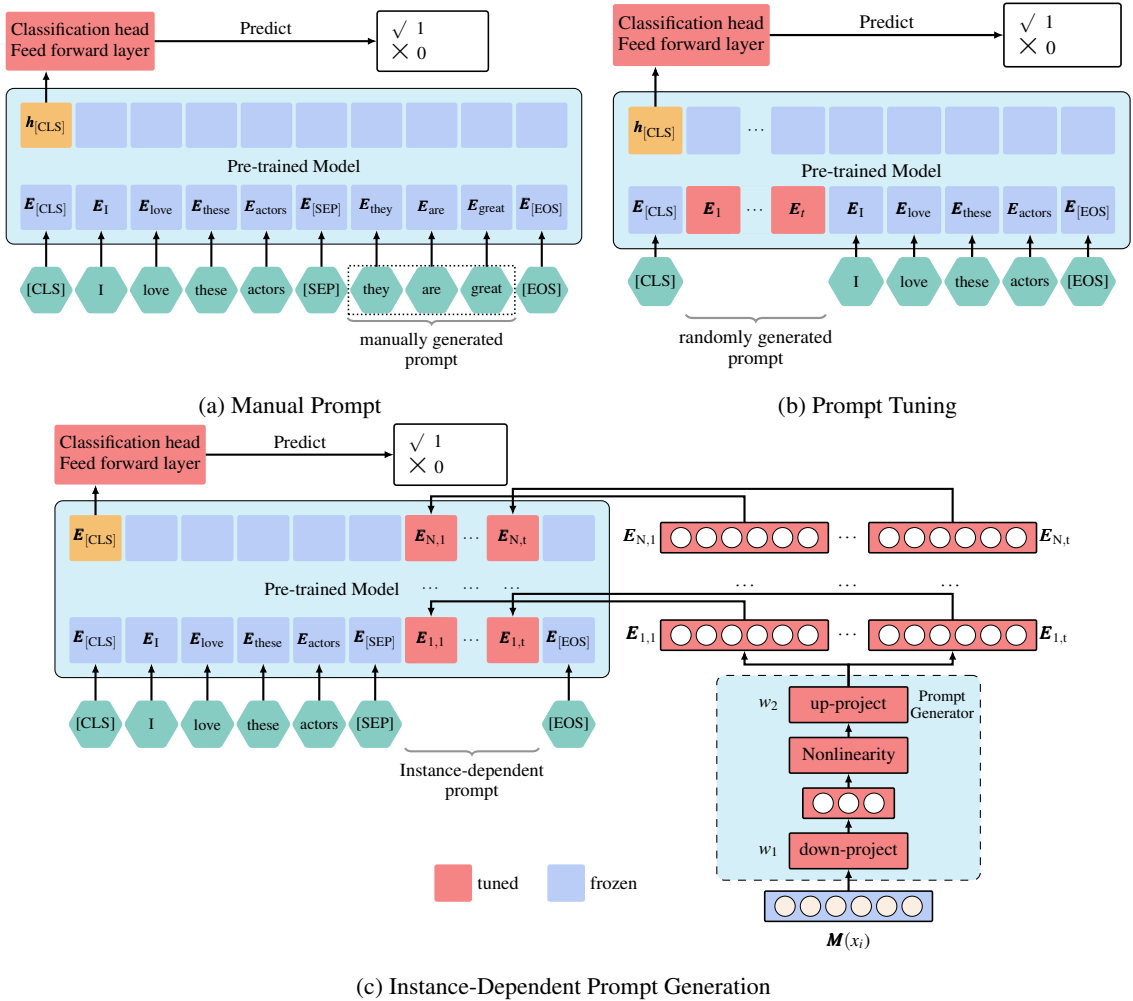
\begin{figure*}[!thbp]
		\centering
		\captionsetup[subfigure]{width=0.48\textwidth}
		\subcaptionbox{Manual Prompt}[.5\linewidth]{
		\resizebox{0.96\linewidth}{!}{
			\begin{tikzpicture}[scale=1, line width=0.6pt, >=stealth,node distance=0.5mm]
				\node[square] (e-CLS) at (0,0){$\bm{E}_{\mathrm{[CLS]}}$};
				\node[square,right=of e-CLS] (e-I) {$\bm{E}_{\mathrm{I}}$};
				\node[square,right=of e-I] (e-love) {$\bm{E}_{\mathrm{love}}$};
				\node[square,right=of e-love] (e-these) {$\bm{E}_{\mathrm{these}}$};
				\node[square,right=of e-these] (e-actors) {$\bm{E}_{\mathrm{actors}}$};
				\node[square,right=of e-actors] (e-SEP) {$\bm{E}_{\mathrm{[SEP]}}$};
				\node[square,right=of e-SEP] (e-they) {$\bm{E}_{\mathrm{they}}$};
				\node[square,right=of e-they] (e-are) {$\bm{E}_{\mathrm{are}}$};
				\node[square,right=of e-are] (e-great) {$\bm{E}_{\mathrm{great}}$};
				\node[square,right=of e-great] (e-EOS) {$\bm{E}_{\mathrm{[EOS]}}$};
				
				\node[origin,below=5mm of e-CLS] (CLS) {[CLS]};
				\node[origin,below=5mm of e-I] (I) {I};
				\node[origin,below=5mm of e-love] (love) {love};
				\node[origin,below=5mm of e-these] (these) {these};
				\node[origin,below=5mm of e-actors] (actors) {actors};
				\node[origin,below=5mm of e-SEP] (SEP) {[SEP]};
				\node[origin,below=5mm of e-they] (they) {they};
				\node[origin,below=5mm of e-are] (are) {are};
				\node[origin,below=5mm of e-great] (great) {great};
				\node[origin,below=5mm of e-EOS] (EOS) {[EOS]};
				
				\draw[larrow] (CLS) -- (e-CLS);
				\draw[larrow] (I) -- (e-I);
				\draw[larrow] (love) -- (e-love);
				\draw[larrow] (these) -- (e-these);
				\draw[larrow] (actors) -- (e-actors);
				\draw[larrow] (SEP) -- (e-SEP);
				\draw[larrow] (they) -- (e-they);
				\draw[larrow] (are) -- (e-are);
				\draw[larrow] (great) -- (e-great);
				\draw[larrow] (EOS) -- (e-EOS);
				
				\node [densely dotted, draw=black, fit={(they) (great)}, inner sep=0.4] (dotted) {};
				\draw[thick,draw=black!42,decorate,decoration={brace,mirror,raise=1pt}] (dotted.south west) --node[shift={(-90:10pt)}]{\scriptsize manually generated} node[shift={(-90:18pt)}]{\scriptsize prompt} (dotted.south east);
				
				\node[square,above=0.45cm of e-CLS,fill=color-h-square] (h-CLS) {$\bm{h}_{\mathrm{[CLS]}}$};
				
				\begin{scope}[on background layer]
					\node[model,fit={(h-CLS) (e-EOS)}] (pre) {};
				\end{scope}
				
				\node[layer,outer sep=0,above=17mm of pre.west,anchor=west] (ent) {Classification head \\ \scriptsize Feed forward layer};
				\draw[larrow] (h-CLS.north) -- (h-CLS.north |- ent.south);
				
				\node[draw,rectangle,align=left,inner xsep=3pt,outer sep=0,minimum width=15mm,minimum height=5mm,font=\scriptsize,right=30mm of ent,rounded corners=1] (surt) {\hspace{-6mm}\,$\surd$ \;\scriptsize 1 \\ {\large \hspace{-6mm}$\times$} \scriptsize 0};
				
				\draw[larrow] (ent) --node[shift={(90:5pt)}]{\scriptsize Predict} (surt);
			\end{tikzpicture}}
		}%
		\subcaptionbox{Prompt Tuning}[.45\linewidth]{
		\resizebox{0.95\linewidth}{!}{
			\begin{tikzpicture}[scale=1, line width=0.6pt, >=stealth,
				node distance=0.5mm]
				\node[square] (e-CLS) at (0,0){$\bm{E}_{\mathrm{[CLS]}}$};
				\node[square,fill=color-layer,right=of e-CLS] (e-1) {$\bm{E}_{1}$};
				\node[square,fill=color-model,right=-1.5mm of e-1] (e-2) {$\cdots$};
				\node[square,fill=color-layer,right=-1.5mm of e-2] (e-3) {$\bm{E}_{t}$};
				\node[square,fill=color-layer,right=of e-CLS] (e-1) {$\bm{E}_{1}$};
				\node[square,right=of e-3] (e-I) {$\bm{E}_{\mathrm{I}}$};
				\node[square,right=of e-I] (e-love) {$\bm{E}_{\mathrm{love}}$};
				\node[square,right=of e-love] (e-these) {$\bm{E}_{\mathrm{these}}$};
				\node[square,right=of e-these] (e-actors) {$\bm{E}_{\mathrm{actors}}$};
				\node[square,right=of e-actors] (e-EOS) {$\bm{E}_{\mathrm{[EOS]}}$};
				
				\node[origin,below=5mm of e-CLS] (CLS) {[CLS]};
				\node[origin,below=5mm of e-I] (I) {I};
				\node[origin,below=5mm of e-love] (love) {love};
				\node[origin,below=5mm of e-these] (these) {these};
				\node[origin,below=5mm of e-actors] (actors) {actors};
				\node[origin,below=5mm of e-EOS] (EOS) {[EOS]};
				
				\draw[larrow] (CLS) -- (e-CLS);
				\draw[larrow] (I) -- (e-I);
				\draw[larrow] (love) -- (e-love);
				\draw[larrow] (these) -- (e-these);
				\draw[larrow] (actors) -- (e-actors);
				\draw[larrow] (EOS) -- (e-EOS);
				
				\node[square,above=0.45cm of e-CLS,fill=color-h-square] (h-CLS) {$\bm{h}_{\mathrm{[CLS]}}$};
				\node[layer,outer sep=0,above=17mm of pre.west,anchor=west] (cla) {Classification head \\ \scriptsize Feed forward layer};
				\draw[larrow] (h-CLS.north) -- (h-CLS.north |- cla.south);
				
				\node[draw,rectangle,align=left,inner xsep=3pt,outer sep=0,minimum width=15mm,minimum height=5mm,font=\scriptsize,right=30mm of ent,rounded corners=1] (surt) {\hspace{-6mm}\,$\surd$ \;\scriptsize 1 \\ {\large \hspace{-6mm}$\times$} \scriptsize 0};
				
				\draw[larrow] (cla) --node[shift={(90:5pt)}]{\scriptsize Predict} (surt);
				
				\begin{scope}[on background layer]
					\node[model,fit={(h-CLS) (e-EOS)}] (pre) {};
					\node[fill=color-ice, fit={(e-1) (e-3)},rounded corners=1,inner sep=0] (e1-3) {};
				\end{scope}
				
				\draw[thick,draw=black!42,decorate,decoration={brace,mirror,raise=1pt}] ([yshift=-10mm]e1-3.south west) --node[shift={(-90:10pt)}]{\scriptsize randomly generated} node[shift={(-90:18pt)}]{\scriptsize prompt} ([yshift=-10mm]e1-3.south east);
				
			\end{tikzpicture}}
		}
		
		\subcaptionbox{Instance-Dependent Prompt Generation}[1\linewidth]{
		\resizebox{0.85\linewidth}{!}{
			\begin{tikzpicture}[scale=1, line width=0.6pt, >=stealth,
				node distance=0.5mm]
				\node[square] (e-CLS) at (0,0){$\bm{E}_{\mathrm{[CLS]}}$};
				\node[square,right=of e-CLS] (e-I) {$\bm{E}_{\mathrm{I}}$};
				\node[square,right=of e-I] (e-love) {$\bm{E}_{\mathrm{love}}$};
				\node[square,right=of e-love] (e-these) {$\bm{E}_{\mathrm{these}}$};
				\node[square,right=of e-these] (e-actors) {$\bm{E}_{\mathrm{actors}}$};
				\node[square,right=of e-actors] (e-EOS) {$\bm{E}_{\mathrm{[EOS]}}$};
				
				\node[origin,below=5mm of e-CLS] (CLS) {[CLS]};
				\node[origin,below=5mm of e-I] (I) {I};
				\node[origin,below=5mm of e-love] (love) {love};
				\node[origin,below=5mm of e-these] (these) {these};
				\node[origin,below=5mm of e-actors] (actors) {actors};
				\node[origin,below=5mm of e-EOS] (EOS) {[EOS]};
				
				\draw[larrow] (CLS) -- (e-CLS);
				\draw[larrow] (I) -- (e-I);
				\draw[larrow] (love) -- (e-love);
				\draw[larrow] (these) -- (e-these);
				\draw[larrow] (actors) -- (e-actors);
				\draw[larrow] (EOS) -- (e-EOS);

				\node[layer,outer sep=0, above=10.0mm of e-I,anchor=west] (pro) {Prompt  Generator};
				
				\begin{scope}[on background layer]
			    	\node[model1,fit={(e-CLS) (EOS)}] (pre) {};
				\end{scope}
				
				\draw[larrow] (pre.north) -- (pre.north |- pro.south);
				
				\node[square] (e-CLS-2) at (6.0,0){$\bm{E}_{\mathrm{[CLS]}}$};
				\node[square,right=of e-CLS-2] (e-I-2) {$\bm{E}_{\mathrm{I}}$};
				\node[square,right=of e-I-2] (e-love-2) {$\bm{E}_{\mathrm{love}}$};
				\node[square,right=of e-love-2] (e-these-2) {$\bm{E}_{\mathrm{these}}$};
				\node[square,right=of e-these-2] (e-actors-2) {$\bm{E}_{\mathrm{actors}}$};
				\node[square,right=of e-actors-2] (e-SEP-2) {$\bm{E}_{\mathrm{[SEP]}}$};
				\node[square,fill=color-layer,right=of e-SEP-2] (e-1-2) {$\bm{E}_{1}$};
				\node[square,fill=color-model,right=-1.5mm of e-1-2] (e-2-2) {$\cdots$};
				\node[square,fill=color-layer,right=-1.5mm of e-2-2] (e-3-2) {$\bm{E}_{t}$};
				\node[square,fill=color-layer,right=of e-SEP-2] (e-1-2) {$\bm{E}_{1}$};
				\node[square,right=of e-3-2] (e-EOS-2) {$\bm{E}_{\mathrm{[EOS]}}$};
				
				\node[origin,below=5mm of e-CLS-2] (CLS-2) {[CLS]};
				\node[origin,below=5mm of e-I-2] (I-2) {I};
				\node[origin,below=5mm of e-love-2] (love-2) {love};
				\node[origin,below=5mm of e-these-2] (these-2) {these};
				\node[origin,below=5mm of e-actors-2] (actors-2) {actors};
				\node[origin,below=5mm of e-SEP-2] (SEP-2) {[SEP]};
				\node[origin,below=5mm of e-EOS-2] (EOS-2) {[EOS]};
				
				\draw[larrow] (CLS-2) -- (e-CLS-2);
				\draw[larrow] (I-2) -- (e-I-2);
				\draw[larrow] (love-2) -- (e-love-2);
				\draw[larrow] (these-2) -- (e-these-2);
				\draw[larrow] (actors-2) -- (e-actors-2);
				\draw[larrow] (SEP-2) -- (e-SEP-2);
				\draw[larrow] (EOS-2) -- (e-EOS-2);
				
				\node[square,above=0.45cm of e-CLS-2,fill=color-h-square] (h-CLS-2) {$\bm{h}_{\mathrm{[CLS]}}$};
				
				\begin{scope}[on background layer]
					\node[model,fit={(h-CLS-2) (e-EOS-2)}] (pre-2) {};
				\end{scope}
				
				\node[layer,outer sep=0,above=17mm of pre-2.west,anchor=west] (ent-2) {Classification head \\ \scriptsize Feed forward layer};
				\draw[larrow] (h-CLS-2.north) -- (h-CLS-2.north |- ent-2.south);
				
				\node[draw,rectangle,align=left,inner xsep=3pt,outer sep=0,minimum width=15mm,minimum height=5mm,font=\scriptsize,right=25mm of ent-2,rounded corners=1] (surt-2) {\hspace{-6mm}\,$\surd$ \;\scriptsize 1 \\ {\large \hspace{-6mm}$\times$} \scriptsize 0};;
				\draw[larrow] (ent-2) --node[shift={(90:5pt)}]{\scriptsize Predict} (surt-2);

				\draw[larrow] ([xshift=3.6mm] pro.north west) -- ++(0,0.2) -- ++(3.7,0) -- ++(0,-3.5) -| (e-1-2);
				\draw[larrow] (pro.north) -- ++(0,0.35) -- ++(3.25,0) -- ++(0,-3.8) -| (e-2-2);
				\draw[larrow] ([xshift=-3.6mm] pro.north east) -- ++(0,0.50) -- ++(2.8,0) -- ++(0,-4.1) -| (e-3-2);
				
				\draw[thick,draw=black!42,decorate,decoration={brace,mirror,raise=1pt}] ([yshift=-17mm]e-1-2.south west) --node[shift={(-90:10pt)}]{\scriptsize Instance-dependent} node[shift={(-90:18pt)}]{\scriptsize prompt} ([yshift=-17mm]e-3-2.south east);
				
				\fill[fill=color-layer,rounded corners=1]  (5.2,-2.7) rectangle node[shift={(0:17pt)}]{\scriptsize tuned} (5.7,-2.2);
				\fill[fill=color-model,rounded corners=1]  (6.8,-2.7) rectangle node[shift={(0:18pt)}]{\scriptsize frozen} (7.3,-2.2);
			\end{tikzpicture}}
		}
		\vspace{-0.1in}
		\caption{An illustration of (a) manual prompt; (b) prompt-tuning method; (c) our proposed method. The red block refers to the trainable module, while the blue block refers to the frozen module.}
		\vspace{-0.2in}
		\label{fig:model}
	\end{figure*}

}
\section{Instance-Dependent Prompt Generation (IDPG)} \label{model}
We now introduce our proposed method, IDPG, along with various model optimizations. The main procedure is illustrated in Figure~\ref{fig:model}.

\subsection{Instance-Dependent Generation} \label{proposed-method}
Let us assume a task $T$ with training data $D_{train} = \{(x_i, y_i)\}_{i=1}^K$. Following prompt tuning, we define the input $x_i = \mathbf{E}(\texttt{[SEP]}S_1\texttt{[SEP]} S_2\texttt{[EOS]})$ for sentence-pair task or $x_i = \mathbf{E}(\texttt{[SEP]}S_1\texttt{[EOS]})$ for single-sentence task, where $\mathbf{E}(\cdot)$ is the token embedding for input sentences. Different from all previous works that only define a task-specific prompt $\mathbf{W}_p(T)\in\mathbb{R}^{d\times t}$, where $t$ is the number of tokens in prompt representation and $d$ is the hidden dimension, we propose a instance-dependent prompt generation method. Specifically, we suppose that the generation of prompt should not only depend on the task $T$, but also be affected by input sequence $x_i$. If $\textbf{M}(x_i)\in\mathbb{R}^d$ is a representation of the input sequence $x_i$ from same pre-trained LM $\textbf{M}$, we design a lightweight model $\mathbf{G}$ to generate the prompt,
\begin{equation}
\mathbf{W}_p(T, x_i)=\mathbf{G}(\mathbf{M}(x_i), T), \; x_i \in D_{train}
\end{equation}
Then, we insert a prompt $\mathbf{W}_p(T)$ together with input sequence $x_i$ to infer $y_i$ during fine-tuning. In this way, we have a unified template 
\begin{equation}
\texttt{softmax}(\textbf{W}\mathbf{h}_{\texttt{[CLS]}})
\end{equation}
\begin{equation}
\mathbf{h}_{\texttt{[CLS]}} = \textbf{M}(\texttt{concat}[x_i, \mathbf{W}_p(T, x_i)])
\end{equation}
where $\mathbf{W}$ is the trainable LM classification head. 

To reduce the number of trainable parameters in $\mathbf{G}$, we apply a lightweight bottleneck architecture (i.e., a two-layer perceptron) for generation. As illustrated in Figure~\ref{fig:model} (c), the generator $\mathbf{G}$ first projects the original $d$-dimensional sentence representation $\mathbf{h}_i$ into $m$ dimensions. After passing through a nonlinear function, generator $\mathbf{G}$ projects the hidden representation back to a $d$ dimensions with $t$ timestamps. The total number of parameters for generator $\mathbf{G}$ is $m(d+1)+td(m+1)$ (bias term included). This model can be regarded as the general version of prompt tuning: the bias term $t\times d$ in the second layer of $\mathbf{G}$ is a task-specific prompt, with preceding parts generating an instance-dependent prompt. The final prompt our method generated is a combination of both. We can control the added number of trainable parameters by setting $m \ll d$, but it is still expensive since hidden dimension $d$ is usually large (1024 in BERT/RoBERTa-Large). In the sequel, we will introduce a parameter squeezing method to further reduce trainable parameters without sacrificing performance. 

Note that our proposed method relies on the input sentence representation $\textbf{M}(x_i)$ to generate prompts. One caveat is that this method will have two forward passes of the pre-trained LM during inference time -- first to generate $\textbf{M}(x_i)$ and then to generate classification results. However, the sentence representation $\textbf{M}(x_i)$ used in our method is task-agnostic. In practice, we can cache the prediction $\textbf{M}(x_i)$ and use it in various downstream tasks or rely on a lightweight sentence representation such as GloVe~\citep{pennington2014glove} (Cf. Section~\ref{4-5-1}).

\subsection{Optimization} \label{multi-position}
We propose two optimization techniques to further improve our proposed method.

\subsubsection{Parameterized Hypercomplex Multiplication (PHM) Layers} Inspired by the recent application of parameterized hypercomplex multiplication (PHM) layers~\cite{zhang2021beyond} in Compacter~\cite{mahabadi2021compacter}, we leverage PHM layers to optimize our prompt generator, $\mathbf{G}$. 
Generally, the PHM layer is a fully-connected layer with form $y = \mathbf{W}x+b$, where $x\in\mathbb{R}^d$ is the input feature, $y\in \mathbb{R}^m$ is the output feature, and $\mathbf{W}\in\mathbb{R}^{m \times d}$ and $b\in\mathbb{R}^m$ are the trainable parameters. When $m$ and $d$ are large, the cost of learning $\mathbf{W}$ becomes the main bottleneck. PHM replaces the matrix $\mathbf{W}$ by a sum of Kronecker products of several small matrices. Given a user-defined hyperparameter $n\in \mathbb{Z}^+$ that divides $m$ and $d$, $\mathbf{W}$ can be calculated as follows:
\begin{equation}\label{eq4}
\mathbf{W}=\sum\limits_{i=1}^{n} \mathbf{A}_i \bigotimes \mathbf{B}_i 
\end{equation}
where $\mathbf{A}_i \in \mathbb{R}^{n\times n}$, $\mathbf{B}_i \in \mathbb{R}^{\frac{m}{n} \times \frac{d}{n}}$, and $\bigotimes$ is Kronecker product. In this way, the number of trainable parameters is reduced to $n \times (n \times n + \frac{m}{n} \times \frac{d}{n}) = n^3 + \frac{m \times d}{n}$. As $n$ is usually much smaller than $m$ and $d$, PHM reduces the amount of parameters by a factor of $n$. 

Suppose that we have a two layer perceptron with down-sample projection $\mathbf{W}_1 \in \mathbb{R}^{m \times d}$ and up-sample projection $\mathbf{W}_2 \in \mathbb{R}^{t \times d \times m}$, where $d$ is the input embedding dimension, $m$ is the hidden layer dimension, and $t$ is the number of tokens we generate. For example, we use RoBERTa-Large with hidden size $d = 1024$, generator hidden size $m=256$, $n = 16$, prompt length $t=5$. By substituting the $\mathbf{W}_1$ and $\mathbf{W}_2$ by two PHM layers and letting $A_i$ shared by both layers, we can reduce the number of parameters from 1.5M to 105K.

\subsubsection{Multi-layer Prompt Tuning} \label{sec-sup}
Prompt tuning~\cite{lester2021power} and P-tuning~\cite{liu2021gpt} both insert continuous prompts into the first transformer layer (cf.~Figure~\ref{fig:model}(b)). While proven efficient in some specific settings, single layer prompt tuning has two main limitations: (i) Capturing deep contextual information: the impact of the first-layer prompts on final prediction is low when transformer goes deeper. (ii) Generalizing to long sequence tasks: it is unclear that prompt tuning can perform well in tasks with long input when only a limited number of parameters can be inserted in single layer.

Following Prefix tuning~\cite{li2021prefix} and P-tuning v2~\cite{liu2021p}, we prepend our generated prompts at each transformer layer to address the above issues. 
However, simply generalizing our model (IDPG) to a multi-layer version (M-IDPG), will significantly increase the number of training parameters, since each layer requires an independent generator $\mathbf{G}$. Instead, we explore different architectures in Section~\ref{4-5-3} to balance the number of tuned parameters against model performance. In short, assuming each layer generator $G_i$ has form $y = \mathbf{W}x+b_i$, we share the weight matrix $\mathbf{W}$ across generators and set the bias term $b_i\in\mathbb{R}^m$ to be layer-specific, where $i = 1,\dots, N$ is the layer index and $N$ is the number of transformer layers.

\section{Experiment Results} \label{experiment}
\subsection{Experimental Setup} \label{setup}
We evaluate on ten standard natural language understanding (NLU) datasets -- MPQA~\cite{wiebe2005annotating}, Subj~\cite{pang2004sentimental}, CR~\cite{hu2004mining}, MR~\cite{pang2005seeing}, and six tasks from GLUE~\citep{wang2018glue}, viz.~SST-2, QNLI, RTE, MRPC, STS-B~\cite{cer2017semeval} and QQP. We compare our proposed method with a wide range of methods, as follows:

\textbf{Transformer fine-tuning:} We instantiated two versions -- a vanilla transformer fine-tuning~\cite{liu2019roberta} and the entailment-based fine-tuning~\cite{wang2021entailment}.

\textbf{Prompt tuning:} We implemented two versions -- standard prompt tuning~\cite{lester2021power} and multi-layer prompt tuning~\cite{li2021prefix, liu2021p}.

\textbf{Adapter-based fine-tuning:} This efficient transfer learning method inserts an adaptation module inside each transformer layer including Compactor~\cite{mahabadi2021compacter} and Adapter~\cite{houlsby2019parameter}.

We compare these against two versions of single-layer instance-dependent generation methods: {S-IDPG-DNN} and {S-IDPG-PHM}. The first version is based on a 2-layer perceptron generator, which contains 1.5M parameters. The second one uses the PHM layer and only contains 105K parameters.

We also explore three versions of multi-layer instance-dependent generation methods: {M-IDPG-DNN}, {M-IDPG-PHM}, {M-IDPG-PHM-GloVe}. Again, the difference between the first two is in the prompt generator, while {M-IDPG-PHM-GloVe} uses GloVe to encode input sequences. 

For a fair comparison, all the pre-trained LMs are 24-layer 16-head RoBERTa-Large models~\cite{liu2019roberta}. 
Additional training details can be found in Appendix~\ref{sec:aexp}. Notably, Prompt-tuning-134 uses 134 prompt lengths in Table~\ref{tab:main}, and it is set so to match the training parameters of the proposed method, M-IDPG-PHM. 
\begin{table*}[t]
\caption{Main results of different transfer learning method. Each methods are evaluated on full test sets (dev sets for GLUE tasks).
We report average results across 5 runs with different initialization. \textbf{Bold} marks the best result among all competing methods. \underline{Underline} marks the best result among all prompt tuning methods. We report the average of accuracy and F1 for both MRPC and QQP, and average of Pearson and Spearman correlation coefficients for STS-B. For all the other tasks, we report accuracy.} 
\label{tab:main}
\centering
\setlength{\tabcolsep}{8pt}
\begin{adjustbox}{max width=0.96\textwidth}
\begin{tabular}{lcccccccccc|c}  %
\toprule
 \textbf{Method} & \textbf{MPQA} & \textbf{Subj} & \textbf{CR} & \textbf{MR} & \textbf{SST-2} & \textbf{QNLI} & \textbf{RTE} & \textbf{MRPC} & \textbf{STS-B} & \textbf{QQP} & Avg\\ 
\midrule
\multicolumn{6}{l}{\emph{Transformer Fine-tuning}} \\
\midrule
RoBERTa & 90.4$_{\pm0.2}$ & 97.1$_{\pm0.1}$ & 90.7$_{\pm 0.7}$ & 91.7$_{\pm0.2}$ & 96.4$_{\pm0.2}$ & \textbf{94.7}$_{\pm 0.1}$ & 85.7$_{\pm0.2}$ & {91.8}$_{\pm0.4}$ & 92.2$_{\pm0.2}$ & \textbf{91.0}$_{\pm0.1}$ & 92.2 \\
EFL & 90.3$_{\pm0.2}$ & 97.2$_{\pm0.1}$ & 93.0$_{\pm 0.7}$ & 91.7$_{\pm0.2}$ & \textbf{96.5}$_{\pm0.1}$ & 94.4$_{\pm 0.1}$ & 85.6$_{\pm2.4}$ & 91.2$_{\pm0.4}$ & \textbf{92.5}$_{\pm0.1}$ & \textbf{91.0}$_{\pm0.2}$ & 92.3 \\
\midrule
\multicolumn{6}{l}{\emph{Adapter}} \\
\midrule
Compacter & 91.1$_{\pm0.2}$ & 97.5$_{\pm0.1}$ & 92.7$_{\pm 0.4}$ & \textbf{92.6}$_{\pm0.2}$ & 96.0$_{\pm0.2}$ & 94.3$_{\pm 0.2}$ & 87.1$_{\pm1.4}$ & 91.6$_{\pm0.6}$ & 91.6$_{\pm0.1}$ & 87.1$_{\pm0.2}$ & 92.2 \\
Adapter & 90.8$_{\pm0.2}$ & 97.5$_{\pm0.1}$ & 92.8$_{\pm 0.3}$ & 92.5$_{\pm0.1}$ & 96.1$_{\pm0.1}$ & 94.8$_{\pm 0.2}$ & 88.1$_{\pm0.4}$ & {91.8}$_{\pm0.6}$ & 92.1$_{\pm0.1}$ & 89.9$_{\pm0.1}$ & \textbf{92.6} \\
\midrule
\multicolumn{6}{l}{\emph{Prompting}} \\
\midrule
Prompt-tuning & 90.3$_{\pm 0.2}$ & 95.5$_{\pm 0.4}$ & 91.2$_{\pm 1.1}$& 91.0$_{\pm 0.2}$ & 94.2$_{\pm 0.3}$ & 86.0$_{\pm 0.3}$ & 87.0$_{\pm 0.4}$ & 84.3$_{\pm 0.3}$ & 87.2$_{\pm 0.2}$ & 81.6$_{\pm 0.1}$ & 88.8 \\
Prompt-tuning-134 & 65.7$_{\pm 19}$ & 95.6$_{\pm 0.2}$ & 86.7$_{\pm 3.6}$& 89.7$_{\pm 0.5}$ & 92.0$_{\pm 0.5}$ & 83.0$_{\pm 1.1}$ & 87.4$_{\pm 0.5}$ & 84.1$_{\pm 0.5}$ & 87.6$_{\pm 0.5}$ & 82.4$_{\pm 0.3}$ & 85.4 \\

Ptuningv2 & 90.4$_{\pm0.3}$ & 96.5$_{\pm0.3}$ & 92.7$_{\pm 0.3}$ & 91.6$_{\pm0.1}$ & 94.4$_{\pm0.2}$ & 92.9$_{\pm0.1}$ & 78.4$_{\pm4.3}$ & 91.4$_{\pm0.4}$ & 89.9$_{\pm0.2}$ & 84.4$_{\pm0.4}$ & 90.3 \\

S-IDPG-PHM & 89.6$_{\pm0.3}$ & 94.4$_{\pm0.3}$ & 90.3$_{\pm0.2}$  & 89.3$_{\pm0.4}$ & 94.7$_{\pm0.2}$ &  90.7$_{\pm 0.3}$ & 89.2$_{\pm0.2}$ & 84.3$_{\pm0.8}$ & 84.7$_{\pm0.9}$ & 82.5$_{\pm0.2}$ & 89.0\\

S-IDPG-DNN &  89.5$_{\pm0.7}$ & 94.9$_{\pm0.4}$ & 89.9$_{\pm1.5}$ & 90.2$_{\pm0.6}$ & 95.1$_{\pm0.2}$ &  90.5$_{\pm0.5}$ & \textbf{\underline{89.4}}$_{\pm 0.4}$ & 83.0$_{\pm0.5}$ & 85.3$_{\pm0.7}$ & 82.7$_{\pm 0.3}$ & 89.1\\

M-IDPG-PHM-GloVe & {90.9}$_{\pm0.2}$ & 97.4$_{\pm0.1}$ & {93.3}$_{\pm0.1}$  & \textbf{\underline{92.6}}$_{\pm0.3}$ & 95.4$_{\pm0.2}$ &  94.4$_{\pm 0.2}$ & 82.1$_{\pm0.6}$ & {92.1}$_{\pm0.4}$ & 91.0$_{\pm0.4}$ & 86.3$_{\pm0.2}$ & 91.6\\
 
M-IDPG-PHM & \textbf{\underline{91.2}}$_{\pm0.2}$ & 97.5$_{\pm0.1}$ & 93.2$_{\pm0.3}$  & \textbf{\underline{92.6}}$_{\pm0.3}$ & \underline{96.0}$_{\pm0.3}$ &  \underline{94.5}$_{\pm 0.1}$ & 83.5$_{\pm0.7}$ & \textbf{\underline{92.3}}$_{\pm0.2}$ & 91.4$_{\pm0.4}$ & 86.2$_{\pm0.1}$ & 91.9\\

M-IDPG-DNN &  \textbf{\underline{91.2}}$_{\pm0.3}$ & \textbf{\underline{97.6}}$_{\pm0.2}$ & \textbf{\underline{93.5}}$_{\pm0.3}$ & \textbf{\underline{92.6}}$_{\pm0.1}$ & {95.9}$_{\pm0.1}$ & \underline{94.5}$_{\pm0.2}$ & 85.5$_{\pm 0.6}$ & {91.8}$_{\pm0.3}$ & \underline{91.5}$_{\pm0.2}$ & \underline{86.9}$_{\pm 0.3}$ & \underline{92.1}\\
\bottomrule
\end{tabular}
\end{adjustbox}
\end{table*}

\subsection{Performance in high-resource scenario} \label{exp}

Table~\ref{tab:main} shows the results of all the methods on full datasets across 10 NLU tasks. We observe that: 
(i)~Our proposed method M-IDPG-PHM consistently outperforms the prompt tuning method and Ptuning v2 by average 3.1pt and 1.6pt, respectively (except on the RTE dataset). 
(ii)~Compared with other efficient transfer learning methods, IDPG performs slightly worse than the Compacter~\citep{mahabadi2021compacter} and Adapter~\citep{houlsby2019parameter}, across the ten tasks. However, the gap is mostly from RTE and QQP. Note that IDPG uses 15K fewer parameters than the Compacter. M-IDPG-PHM is better than Compacter on four tasks and has the same performance on three tasks. 
(iii)~The improvement of our method is more prominent in the single-sentence classification task. The four best results (MPQA, Subj, CR, MR) among all competing methods in single-sentence classification tasks are made by IDPG models.
Specifically, M-IDPG-PHM performs 0.84pt and 0.36pt better than RoBERTa and EFL, respectively. 
(iv)~PHM-based generator performs on par with the DNN-based generator while having a significantly lower number of trainable parameters.
(v)~GloVe-based sentence encoder also performs similar to LM-based sentence encoder, indicating the advancement of instance-dependent prompt generation does not rely on a robust contextual sentence encoder. 
(vi)~When we fix the training parameters to be the same, the comparison between Prompt-tuning-134 and M-IDPG-PHM illustrates that our approach works better than prompt tuning not just because of using more parameters.

\subsection{Efficiency} 

Table~\ref{tab:efficiency} lists the number of trainable parameters for different methods excluding the classification head. The general goal for efficient transfer learning is to train models with fewer parameters while achieving better performance. %
Traditional prompt-tuning method only requires training a token embedding table with a few thousand parameters. However, its performance is worse than a lightweight adapter model (e.g., Compacter with 149K parameters). Our proposed method, especially the M-IDPG-PHM, falls in the gap between prompt-tuning and adapter model, since it only requires training 134K parameters and performs on par with Compacter. 

\begin{table}[h]
\footnotesize
\centering
\resizebox{0.96\columnwidth}{!}{
\begin{tabular}{l|r}
    \toprule
    \textbf{Method} & \textbf{\# Parameters}\\
    \midrule
    Transformer Fine-tune~\citep{liu2019roberta} & 355M\\
    Adapter~\citep{houlsby2019parameter} & 1.55M \\
    Compacter~\citep{mahabadi2021compacter} & 149K \\
    Prompt-tuning~\citep{lester2021power} & 5K \\
    Prompt-tuning-134~\citep{lester2021power} & 134K \\
    P-Tuningv2~\cite{liu2021p} & 120K \\
    S-IDPG-PHM & 105K \\
    S-IDPG-DNN & 1.5M \\
    M-IDPG-PHM-GloVe & 141K \\
    M-IDPG-PHM & 134K \\
    M-IDPG-DNN & 216K \\
    \bottomrule
\end{tabular}
}
\caption{
    Number of trainable parameters of different methods. Note that we did not include the parameters from classification heads.
}
\label{tab:efficiency}
\end{table}

\begin{table*}[t]
\caption{Low-resource results are evaluated on full test sets. We report average results across 5 runs with different initialization. \textbf{Bold} marks the best result among all competing methods. \underline{Underline} marks the best result among all prompt tuning methods. We report the average of accuracy and F1 for both MRPC and QQP, and average of Pearson and Spearman correlation coefficients for STS-B. For all other tasks, we report accuracy.} 
\label{tab:few-single}
\centering
\setlength{\tabcolsep}{8pt}
\begin{adjustbox}{max width=0.96\textwidth}
\begin{tabular}{lcccccccccc|c}  %
\toprule
 \textbf{Method} & \textbf{MPQA} & \textbf{Subj} & \textbf{CR} & \textbf{MR} & \textbf{SST-2} & \textbf{QNLI} & \textbf{RTE} & \textbf{MRPC} & \textbf{STS-B} & \textbf{QQP} &  Avg\\ 
\midrule
$K$ = 100 & & & & & & & &\\
\midrule
Fine-tuning &  \textbf{86.2}$_{\pm 0.4}$ & {88.4}$_{\pm 0.8}$ & 83.7$_{\pm 2.4}$ & 81.4$_{\pm 1.0}$ & {86.2}$_{\pm 1.3}$ & 77.7$_{\pm1.5}$ & 84.2$_{\pm 1.2}$ & 72.6$_{\pm 3.7}$ & \textbf{84.1}$_{\pm 1.6}$& \textbf{78.1}$_{\pm 0.4}$ & \textbf{82.2}  \\
Adapter-tuning & 81.0$_{\pm 2.9}$ & {88.7}$_{\pm 0.8}$ & \textbf{84.7}$_{\pm 2.1}$ & \textbf{83.7}$_{\pm 0.7}$ & {85.7}$_{\pm 0.9}$ & 75.6$_{\pm0.8}$ & 84.7$_{\pm 0.6}$ & 80.0$_{\pm 0.9}$ & {78.1}$_{\pm 1.4}$& 77.1$_{\pm 0.6}$ & {81.9}  \\
\cdashline{1-12}
prompt tuning & 75.9$_{\pm 1.6}$ & 86.8$_{\pm 0.8}$ & 72.9$_{\pm 1.4}$ & 74.1$_{\pm 1.4}$ & 82.9$_{\pm 2.0}$ & \textbf{\underline{82.7}}$_{\pm0.2}$& 86.5$_{\pm0.6}$ & {80.0}$_{\pm 1.3}$ & 70.2$_{\pm 3.1}$& {76.5}$_{\pm 0.4}$ & 78.9 \\ 
 P-Tuningv2 &  {74.3}$_{\pm 2.9}$ & {89.7}$_{\pm 0.8}$ & {80.1}$_{\pm 1.0}$ & 82.5$_{\pm 1.1}$ & {85.1}$_{\pm 1.6}$ & 78.2$_{\pm0.5}$ & 83.6$_{\pm 0.7}$ & \textbf{\underline{80.1}}$_{\pm 0.6}$ & {78.8}$_{\pm 3.0}$& 76.8$_{\pm 0.5}$ & 80.9  \\
S-IDPG-PHM &  \underline{79.0}$_{\pm 3.7}$ & {87.6}$_{\pm 1.1}$ & 75.0$_{\pm 1.6}$ & 76.2$_{\pm 1.3}$ & {87.6}$_{\pm 1.3}$ & 80.4$_{\pm1.2}$ & 86.3$_{\pm 0.5}$ & 79.3$_{\pm 0.4}$ & {70.9}$_{\pm 2.5}$& 76.1$_{\pm 0.6}$ & {79.8}  \\
S-IDPG-DNN &  78.0$_{\pm 2.1}$ & 84.2$_{\pm 1.6}$ & {76.3}$_{\pm 4.5}$ & {77.4}$_{\pm 0.5}$ & \textbf{\underline{89.6}}$_{\pm 1.2}$ & 81.1$_{\pm0.8}$& \textbf{\underline{87.4}}$_{\pm0.8}$ & 78.8$_{\pm 1.3}$ & 70.6$_{\pm 2.8}$ & 74.1$_{\pm 0.9}$ &  {79.8}  \\
M-IDPG-PHM-GloVe &  {76.6}$_{\pm 2.0}$ & \textbf{\underline{90.7}}$_{\pm 0.4}$ & \underline{80.6}$_{\pm 2.6}$ & \underline{83.0}$_{\pm 1.5}$ & {85.6}$_{\pm 0.8}$ & 77.9$_{\pm1.3}$ & 84.4$_{\pm 0.9}$ & 79.6$_{\pm 0.9}$ & {77.8}$_{\pm 1.6}$& 76.1$_{\pm 0.7}$ & {81.2}  \\
M-IDPG-PHM &  {75.5}$_{\pm 4.6}$ & {90.5}$_{\pm 0.6}$ & 80.2$_{\pm 1.5}$ & 82.5$_{\pm 1.1}$ & {85.9}$_{\pm 1.2}$ & 78.8$_{\pm1.6}$ & 84.0$_{\pm 0.4}$ & 79.9$_{\pm 0.8}$ & \underline{79.3}$_{\pm 0.4}$& \underline{77.1}$_{\pm 0.2}$ & \underline{81.4}  \\
\midrule
$K$ = 500 & & & & & & \\
\midrule
Fine-tuning &  85.1$_{\pm 1.7}$ & {94.1}$_{\pm 0.4}$ & \textbf{90.9}$_{\pm 0.6}$ & {87.6}$_{\pm 0.5}$ & \textbf{92.5}$_{\pm 0.6}$ & 85.7$_{\pm 0.6}$ & 57.5$_{\pm 1.0}$ & {82.3}$_{\pm 0.6}$ & \textbf{88.8}$_{\pm 0.5}$ & \textbf{79.0}$_{\pm 0.3}$ & 84.3 \\
Adapter-tuning &  \textbf{86.0}$_{\pm 0.8}$ & {94.9}$_{\pm 0.2}$ & 89.5$_{\pm 1.0}$ & {88.5}$_{\pm 0.2}$ & 91.9$_{\pm 0.9}$ & 82.2$_{\pm 0.6}$ & 83.9$_{\pm 0.8}$ & {82.7}$_{\pm 0.5}$ & 86.6$_{\pm 0.5}$ & {78.9}$_{\pm 0.3}$ & \textbf{86.5} \\
\cdashline{1-12}
prompt tuning &  82.4$_{\pm 1.3}$ & 91.2$_{\pm 0.1}$ & {86.8}$_{\pm 0.4}$ & 84.6$_{\pm 0.8}$ & {88.6}$_{\pm 1.0}$ & \textbf{\underline{86.3}}$_{\pm0.4}$ & 86.5$_{\pm0.4}$ & 80.0$_{\pm 0.4}$ & 77.4$_{\pm 1.9}$ & 77.8$_{\pm 0.3}$ & 84.2  \\ 
P-Tuningv2 &  84.0$_{\pm 1.3}$ & {94.6}$_{\pm 0.3}$ & 89.0$_{\pm 1.8}$ & {88.1}$_{\pm 0.5}$ & 91.3$_{\pm 0.7}$ & 84.6$_{\pm 0.8}$ & 84.2$_{\pm 1.5}$ & \textbf{\underline{83.2}}$_{\pm 0.7}$ & 83.8$_{\pm 0.5}$ & \underline{78.6}$_{\pm 0.3}$ & 86.1 \\
S-IDPG-PHM &  81.6$_{\pm 2.7}$ & {91.4}$_{\pm 0.7}$ & 85.8$_{\pm 2.0}$ & {85.8}$_{\pm 0.5}$ & 88.5$_{\pm 1.3}$ & 85.0$_{\pm 0.4}$ & 86.3$_{\pm 1.3}$ & {81.9}$_{\pm 0.8}$ & 78.3$_{\pm 1.5}$ & {78.1}$_{\pm 0.3}$ & 84.3 \\
S-IDPG-DNN & {84.8}$_{\pm 0.7}$ & 90.8$_{\pm 0.6}$ & \underline{89.7}$_{\pm 1.0}$ & {86.1}$_{\pm 2.8}$ & {90.4}$_{\pm 1.6}$ & 84.8$_{\pm 0.3}$ & \textbf{\underline{87.7}}$_{\pm 0.7}$ & {82.0}$_{\pm 1.4}$ & {79.1}$_{\pm 2.3}$ & 77.1$_{\pm 0.4}$ & {85.3} \\
M-IDPG-PHM-GloVe &  84.0$_{\pm 1.7}$ & \textbf{\underline{95.0}}$_{\pm 0.2}$ & 89.0$_{\pm 1.1}$ & {88.1}$_{\pm 0.5}$ & 90.4$_{\pm 1.3}$ & 85.1$_{\pm 0.1}$ & 84.0$_{\pm 1.0}$ & {82.3}$_{\pm 0.5}$ & 84.1$_{\pm 0.8}$ & {78.2}$_{\pm 0.8}$ & 86.0 \\
M-IDPG-PHM &  \underline{85.2}$_{\pm 1.1}$ & {94.6}$_{\pm 0.0}$ & 89.1$_{\pm 1.6}$ & \textbf{\underline{88.8}}$_{\pm 0.4}$ & \underline{91.6}$_{\pm 1.1}$ & 84.9$_{\pm 0.9}$ & 83.9$_{\pm 0.7}$ & {82.5}$_{\pm 0.5}$ & \underline{84.2}$_{\pm 0.5}$ & \underline{78.6}$_{\pm 0.3}$ & \underline{86.3} \\
\midrule
$K$ = 1000 & & & & & & \\
\midrule
Fine-tuning &  87.7$_{\pm 0.7}$ & {95.1}$_{\pm 0.2}$ & {89.8}$_{\pm 1.2}$ & {89.2}$_{\pm 0.5}$ & 93.6$_{\pm 0.4}$ & \textbf{88.0}$_{\pm 0.7}$ & 87.3$_{\pm 1.3}$ & \textbf{87.9}$_{\pm 0.9}$ & \textbf{90.8}$_{\pm 0.2}$ & {79.8}$_{\pm 0.3}$ & \textbf{88.9} \\
Adapter-tuning &  \textbf{88.2}$_{\pm 0.6}$ & {95.6}$_{\pm 0.3}$ & \textbf{89.9}$_{\pm 1.4}$ & \textbf{90.0}$_{\pm 0.3}$ & 92.9$_{\pm 0.2}$ & 85.2$_{\pm 0.7}$ & 86.8$_{\pm 0.7}$ & {86.1}$_{\pm 0.6}$ & 89.6$_{\pm 0.5}$ & \textbf{79.9}$_{\pm 0.3}$ & 88.4 \\
\cdashline{1-12}
prompt tuning &  83.9$_{\pm 2.0}$ & 92.6$_{\pm 0.4}$  & 87.2$_{\pm 1.4}$ & 86.7$_{\pm 0.3}$ & 89.9$_{\pm 1.0}$ & {86.9}$_{\pm 0.1}$ & 86.4$_{\pm 0.7}$ & 82.5$_{\pm 0.3}$  & 82.9$_{\pm 1.3}$ & 78.6$_{\pm 0.3}$ & 85.8 \\ 
P-Tuningv2 &  87.0$_{\pm 0.9}$ & \textbf{\underline{95.9}}$_{\pm 0.4}$ & {88.3}$_{\pm 1.5}$ & {89.5}$_{\pm 0.3}$ & 93.2$_{\pm 0.5}$ & 87.4$_{\pm 0.4}$ & 85.1$_{\pm 1.1}$ & {82.6}$_{\pm 1.1}$ & \underline{87.8}$_{\pm 0.3}$ & \underline{79.3}$_{\pm 0.4}$ & 87.6 \\
S-IDPG-PHM &  83.4$_{\pm 1.7}$ & {93.4}$_{\pm 0.9}$ & {89.2}$_{\pm 0.8}$ & {88.0}$_{\pm 0.9}$ & 90.2$_{\pm 1.0}$ & 85.5$_{\pm 0.6}$ & 86.9$_{\pm 0.6}$ & \underline{83.1}$_{\pm 0.4}$ & 83.9$_{\pm 0.8}$ & {78.9}$_{\pm 0.4}$ &86.3 \\
S-IDPG-DNN & {85.9}$_{\pm 0.8}$ & 93.3$_{\pm 1.2}$ & \textbf{\underline{89.9}}$_{\pm 0.8}$ & {89.6}$_{\pm 1.1}$ & {92.2}$_{\pm 0.8}$ &  85.2$_{\pm1.3}$& \textbf{\underline{87.7}}$_{\pm0.8}$ & 82.5$_{\pm 0.9}$ & {84.7}$_{\pm 0.9}$ & 78.0$_{\pm 0.8}$ &  {86.9} \\
M-IDPG-PHM-GloVe &  86.5$_{\pm 0.7}$ & {95.5}$_{\pm 0.3}$ & {87.7}$_{\pm 1.3}$ & {89.3}$_{\pm 0.4}$ & 93.4$_{\pm 0.3}$ & \underline{87.5}$_{\pm 0.3}$ & 84.9$_{\pm 0.9}$ & {82.7}$_{\pm 0.7}$ & 87.6$_{\pm 0.3}$ & {79.1}$_{\pm 0.7}$ & 87.4 \\
M-IDPG-PHM &  \underline{87.7}$_{\pm 0.5}$ & {95.6}$_{\pm 0.2}$ & {89.2}$_{\pm 1.2}$ & \underline{89.8}$_{\pm 0.4}$ & \textbf{\underline{93.7}}$_{\pm 0.6}$ & 87.2$_{\pm 0.5}$ & 85.6$_{\pm 0.6}$ & {82.5}$_{\pm 0.9}$ & \underline{87.8}$_{\pm 0.8}$ & {79.1}$_{\pm 0.4}$ & \underline{87.8} \\
\bottomrule
\end{tabular}
\end{adjustbox}
\end{table*}

\subsection{Performance in low-resource scenario} \label{exp-few}
We further evaluate our proposed method in the low-resource scenario. Following the existing evaluation protocols in the few-shot setting~\cite{he2021effectiveness}, we sample a subset of the training data for each task with size $K\in\{100, 500, 1000\}$ as our training data and another subset with size $1000$ as a development set. We compare our proposed methods with all prompt tuning methods, one fine-tuning model (EFL), and one adapter tuning model (Compacter). 

In the extreme low-resource case when $K$=100, M-IDPG-PHM performs 2.5pt better than the traditional prompt tuning method and 0.5pt better than the multi-layer P-Tuning v2 method. This improvement illustrates that our method has better generalization in few-shot settings. When $K$ becomes larger, IDPG-PHM still maintains good results with 1.9pt, 0.2pt ($K$=500) and 2.0pt, 0.2pt ($K$=1000) better accuracy than traditional prompt tuning, P-tuning v2, respectively. We also observe that when $K$ is small, our method sometimes has a high variance (e.g., 4.6 on MPQA when $K=100$). We suspect that this may be due to bad initialization that leads the model to non-optimal parameters.

\subsection{Intrinsic Study}
We conduct several ablation studies including exploration of different generator architectures and impact of selecting different prompt positions.

\subsubsection{Sentence Encoder: GloVe or LMs?}\label{4-5-1}
The proposed IDPG method relies on pre-trained LM to extract sentence representation, i.e., \texttt{[CLS]} token embedding. Obtaining contextualized transformer sentence embedding is often expensive if it is not pre-computed. One open question is to explore reliability on lightweight sentence representations such as GloVe embedding~\citep{pennington2014glove} or token embedding of pre-trained language models. 

To answer this question, we apply the pre-trained GloVe word vectors\footnote{Obtained from \url{https://nlp.stanford.edu/data/glove.6B.zip} version glove.6b.300d.txt} to extract the sentence representation. Specifically, we take the average of word vectors as the sentence embeddings:
\begin{equation}
\textbf{M}(x_i)=\frac{1}{k} \sum_{j=1}^{k} \texttt{GloVe}(t_j), \; x_i \in D_{train}
\end{equation}
 where $x_i$ is the input sequence with $k$ tokens $t_1, \dots, t_k$. According to Table~\ref{tab:main}, using GloVe as sentence encoder to generate prompts doesn't sacrifice much performance over the ten tasks and outperforms prompt tuning and P-tuning v2. It indicates that our model does not benefit a lot from a strong contextual pre-trained LM. Instead, a light sentence encoder such as GloVe can also help the tasks. Also, instance-dependent prompt tuning shows promising improvement over non-instance-dependent prompt tuning models. 
 
\subsubsection{Prompt Generator: PHM or DNN?}\label{4-5-2}
To reduce the tuning parameters, we substitute the DNN layers with PHM layers. An open question we seek to answer is what is the best generation model for prompt regardless of training parameters. Hence, we compare the PHM-based prompt generator with the DNN-based prompt generator, as shown in Table~\ref{tab:main}. We observe that including DNN as a generator doesn't improve performance significantly, with +0.1pt gain on average, while adding 87K parameters (with hidden size m=16). On the other hand, this ablation study further verifies PHM layers' efficiency in the generation model. 

\begin{figure}[t]
	\centering
	\begin{center}
		\includegraphics[width=0.45\textwidth]{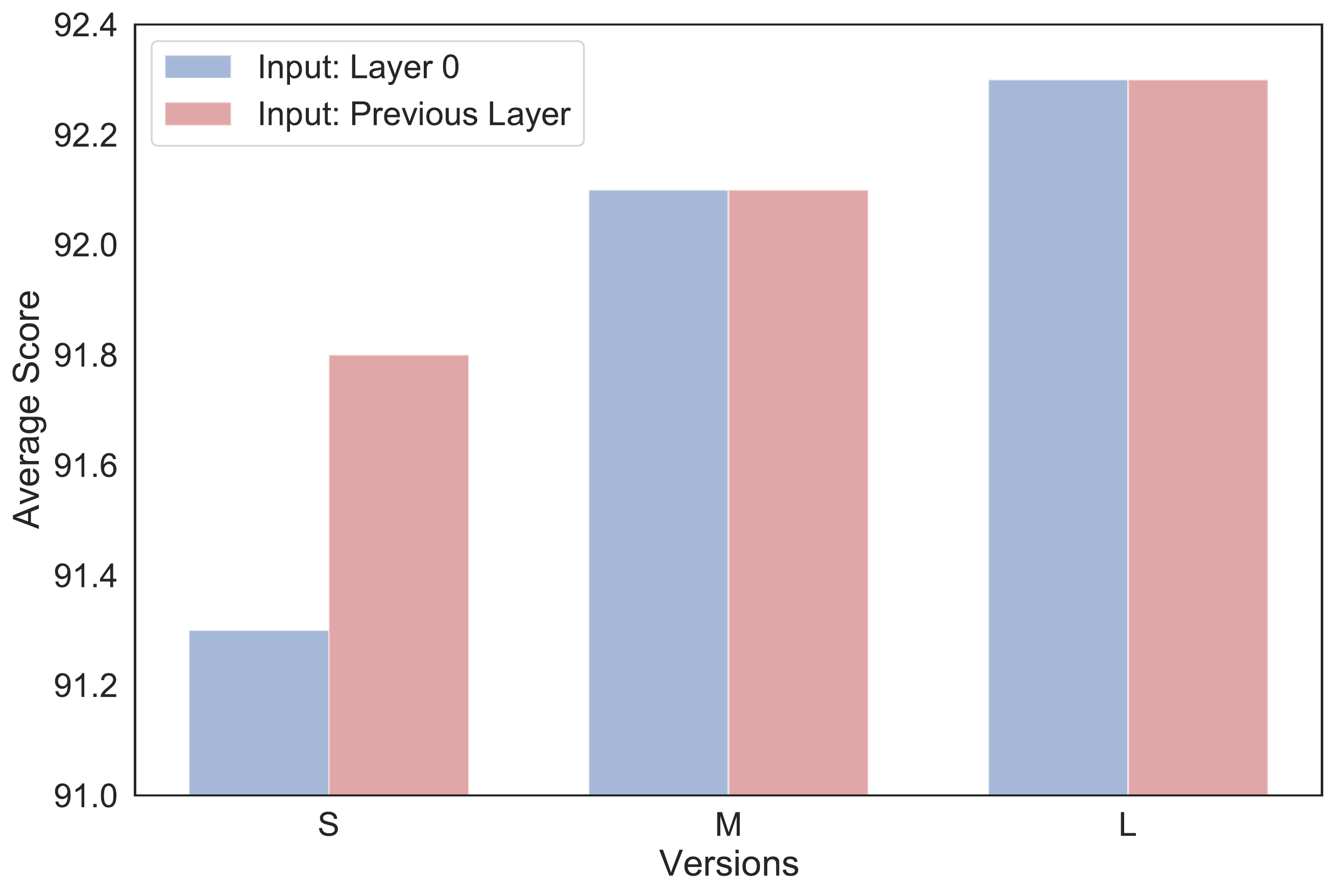}
		\caption[7.5pt]{Comparison between three different multi-layer generator models (S, M, L versions), and comparison between taking layer 0's output or previous layer's output as input. }
		\label{fig:midpg}
		\vspace{-0.2in}
	\end{center}
\end{figure}
\subsubsection{Multi-layer Architecture Exploration}\label{4-5-3}
When applying the instance-dependent generation model $\mathbf{G}$ into a multi-layer case, the first challenge we face is the considerable increase in training parameters. If each transformer layer requires an independent generator $\mathbf{G}_i$, the number of training parameters increases $N$ times, where $N$ is the number of transformer layers (24 in RoBERTa-Large). Assuming $\mathbf{G}$ has the form $y=\textbf{W}x+b$, there are three alternatives: (i) Smallest version (S version): sharing both $W$ and $b$; (ii) Middle version (M version): sharing $W$ and making $b$ layer-specific; and (iii) Largest version (L version): making both $W$ and $b$ layer-specific. 

Another way to reduce the training parameters is by adjusting the hidden size $m$ of the generator. We compare two models with $m=16$ and $m=256$. Surprisingly, we find that generator with a hidden size $16$ is not far from the large model (92.0 vs. 92.1, respectively, in M version). We hypothesize that the smaller hidden size of $16$ is already enough to store useful instance information, and setting $m$ too large may be less efficient. 

Besides, in single-layer prompt generation model, the input to \textbf{G} is $M(x_i)$ - the representation of input sequence $x_i$. In a multi-layer case, the input to each layer generator has another option, i.e., the previous layer's output. However, as shown in Figure~\ref{fig:midpg}, the experiment results suggest no significant difference between the two input ways. As for the generator selection, the three models perform as expected (S version < M version < L version). In Table~\ref{tab:main}, M-IDPG-PHM uses the previous layer's output as input, M version as the generator, and $16$ as the generator hidden size. Detailed information for all models' performance on each task can be found in Appendix~\ref{sec:A3}. 

\subsubsection{Prompt Insertion: Single-layer or Multi-layer?}
P-tuning v2~\cite{liu2021p} conducted substantial ablation studies on the influence of inserting prompt into different transformer layers. To boost single-layer IDPG performance, we add supplementary training (cf.~Appendix~\ref{sec-sup}) and conduct ablation studies in Appendix~\ref{appendix-ablation}. We come to a similar conclusion that multi-layer instance-dependent prompt tuning model (M-IDPG) is significantly better than the single-layer method (S-IDPG) in both evaluation settings. An interesting finding is that the impact of supplementary training on S-IDPG is high while it is limited for M-IDPG. 

\begin{figure}[t]
	\centering
	\begin{center}
		\includegraphics[width=0.45\textwidth]{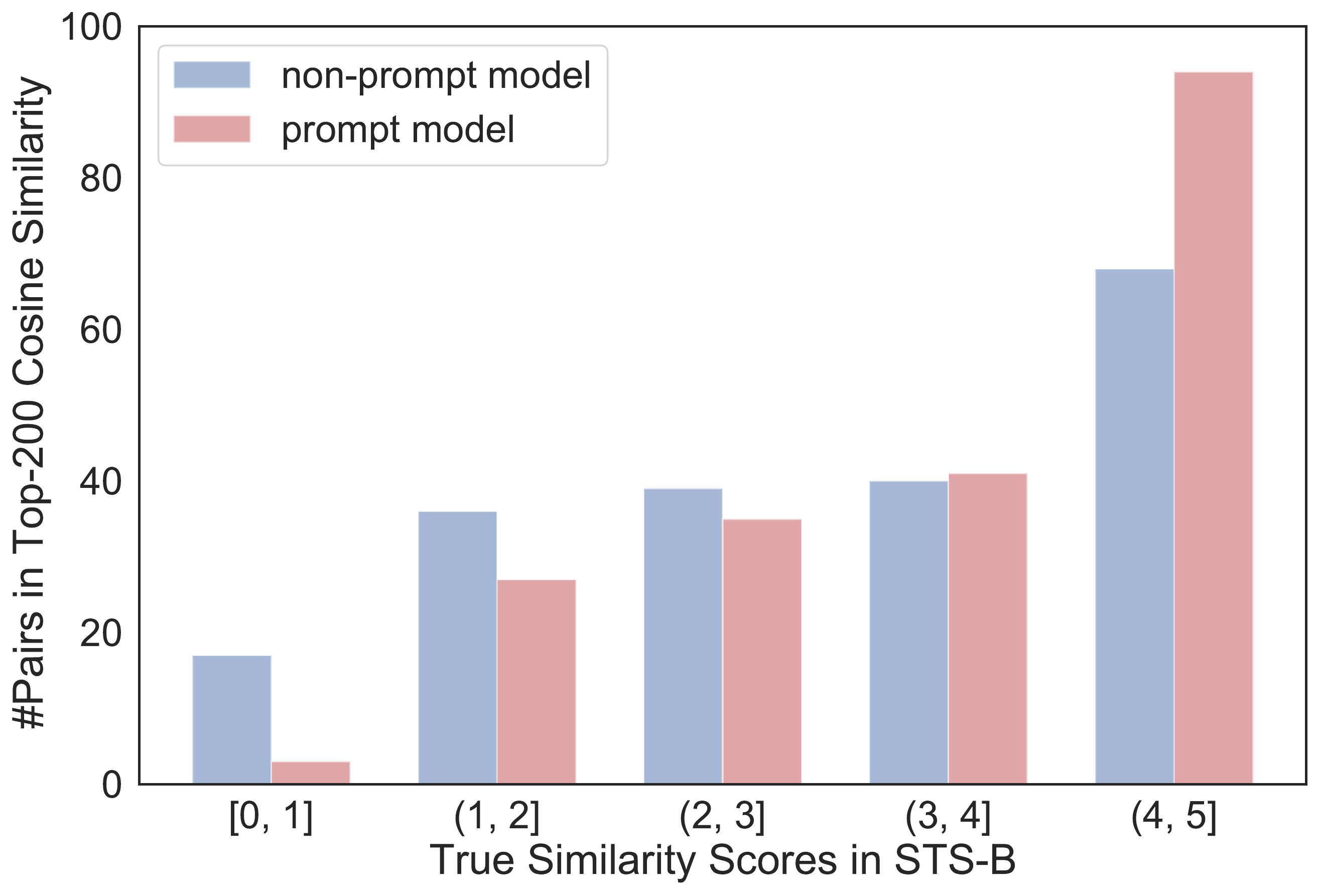}
		\caption[7.5pt]{The number of pairs of each group in Top-200 cosine similarity ranking. More results can be found in Appendix~\ref{appendix-cos}.}
		\label{fig:ana}
		\vspace{-0.2in}
	\end{center}
\end{figure}
\subsubsection{How Prompts Help?}
Given two sentences, we encode each of them by one of the comparison models and compute the cosine similarity. We sort all sentence pairs in STS-B dev set in descending order by the cosine similarity scores and get a distribution for number of pairs in each group that is included in Top-k ranking. We compare a vanilla model without any prompts with M-IDPG-PHM. Both models are fine-tuned on STS-B training set. As shown in Figure~\ref{fig:ana}, prompts bring the similar sentences closer while pushing the dissimilar ones apart.

\subsubsection{IDPG Scalability}
We study our proposed model's scalability in this section. In general, the performance of IDPG in downstream tasks improves gradually when using a larger prompt length (Cf. Appendix~\ref{appendix-len}).

\section{Related Work} \label{related-work}
\noindent \textbf{Supplementary Training:} %
Existing works~\cite{phang2018sentence,liu2019roberta} have observed that starting from the fine-tuned MNLI model results in a better performance than directly from the vanilla pre-trained models for RTE, STS, and MRPC tasks. A series of work (SentenceBERT~\cite{reimers2019sentence}, BERT-flow~\cite{li2020sentence}, SimCSE~\cite{gao2021simcse}) explored intermediate training to improve STS tasks. All of them applied pre-fine tuning on NLI datasets. More recently, EFL~\cite{wang2021entailment} proposed a task transformation paradigm, improving single sentence tasks with less labels using rich sentence-pair datasets. 

\textbf{Adapter Tuning:} Adapter tuning has emerged as a novel parameter-efficient transfer learning paradigm~\cite{houlsby2019parameter, pfeiffer2020mad}, in which adapter layers -- small bottleneck layers -- are inserted and trained between frozen pre-trained transformer layers. On the GLUE benchmark, adapters attain within $0.4\%$ of the performance of full fine-tuning by only training $3.6\%$ parameters per task. %
Compactor~\cite{mahabadi2021compacter} substitutes the down-projector and up-projector matrices by a sum of Kronecker products, reducing the parameters by a large margin while maintaining the overall performance. %

\noindent \textbf{Prompting:} 
Hand-crafted prompts were shown to be helpful to adapt generation in GPT-3~\cite{brown2020language}. Existing works including LM-BFF~\cite{gao2020making, wang2021entailment} explored the prompt searching in a few-shot setting. 

Recently, several researchers have proposed continuous prompts training to overcome the challenges in discrete prompt searching. %
Prefix tuning~\cite{li2021prefix} and P-tuningv2~\cite{liu2021p} prepend a sequence of trainable embeddings at each transformer layer and optimizes them. Two contemporaneous works -- prompt tuning~\cite{lester2021power} and P-tuning~\cite{liu2021gpt}, interleave the training parameters in the input embedding layer instead of each transformer layer. %
All these methods focus on task-specific prompt optimization. Our proposed method, IDPG, is the first prompt generator that is not only task-specific but also instance-specific.

\section{Conclusion and Discussion} \label{conclusion}

We have introduced IDPG, an instance-dependent prompt generation model that generalizes better than the existing prompt tuning methods. Our method first factors in an instance-dependent prompt, which is robust to data variance. Parameterized Hypercomplex Multiplication (PHM) is applied to shrink the training parameters in our prompt generator, which helps us build an extreme lightweight generation model. Despite adding fewer parameters than prompt tuning, IDPG shows consistent improvement. It is also on par with the lightweight adapter tuning methods such as Compacter while using a similar amount of trainable parameters. This work provided a new research angle for prompt-tuning of a pre-trained language model.

\bibliography{main_NAACL2022.bib}

\begin{thebibliography}{31}
\expandafter\ifx\csname natexlab\endcsname\relax\def\natexlab#1{#1}\fi

\bibitem[{Ba et~al.(2016)Ba, Kiros, and Hinton}]{ba2016layer}
Jimmy~Lei Ba, Jamie~Ryan Kiros, and Geoffrey~E Hinton. 2016.
\newblock Layer normalization.
\newblock \emph{arXiv preprint arXiv:1607.06450}.

\bibitem[{Brown et~al.(2020)Brown, Mann, Ryder, Subbiah, Kaplan, Dhariwal,
  Neelakantan, Shyam, Sastry, Askell et~al.}]{brown2020language}
Tom~B Brown, Benjamin Mann, Nick Ryder, Melanie Subbiah, Jared Kaplan, Prafulla
  Dhariwal, Arvind Neelakantan, Pranav Shyam, Girish Sastry, Amanda Askell,
  et~al. 2020.
\newblock Language models are few-shot learners.
\newblock \emph{arXiv preprint arXiv:2005.14165}.

\bibitem[{Cer et~al.(2017)Cer, Diab, Agirre, Lopez-Gazpio, and
  Specia}]{cer2017semeval}
Daniel Cer, Mona Diab, Eneko Agirre, Inigo Lopez-Gazpio, and Lucia Specia.
  2017.
\newblock Semeval-2017 task 1: Semantic textual similarity-multilingual and
  cross-lingual focused evaluation.
\newblock \emph{arXiv preprint arXiv:1708.00055}.

\bibitem[{Devlin et~al.(2019)Devlin, Chang, Lee, and
  Toutanova}]{devlin2019bert}
Jacob Devlin, Ming-Wei Chang, Kenton Lee, and Kristina Toutanova. 2019.
\newblock Bert: Pre-training of deep bidirectional transformers for language
  understanding.
\newblock In \emph{Proceedings of the 2019 Conference of the North American
  Chapter of the Association for Computational Linguistics: Human Language
  Technologies, Volume 1 (Long and Short Papers)}, pages 4171--4186.

\bibitem[{Gao et~al.(2020)Gao, Fisch, and Chen}]{gao2020making}
Tianyu Gao, Adam Fisch, and Danqi Chen. 2020.
\newblock Making pre-trained language models better few-shot learners.
\newblock \emph{arXiv preprint arXiv:2012.15723}.

\bibitem[{Gao et~al.(2021)Gao, Yao, and Chen}]{gao2021simcse}
Tianyu Gao, Xingcheng Yao, and Danqi Chen. 2021.
\newblock Simcse: Simple contrastive learning of sentence embeddings.
\newblock \emph{arXiv preprint arXiv:2104.08821}.

\bibitem[{He et~al.(2016)He, Zhang, Ren, and Sun}]{he2016deep}
Kaiming He, Xiangyu Zhang, Shaoqing Ren, and Jian Sun. 2016.
\newblock Deep residual learning for image recognition.
\newblock In \emph{Proceedings of the IEEE conference on computer vision and
  pattern recognition}, pages 770--778.

\bibitem[{He et~al.(2021)He, Liu, Ye, Tan, Ding, Cheng, Low, Bing, and
  Si}]{he2021effectiveness}
Ruidan He, Linlin Liu, Hai Ye, Qingyu Tan, Bosheng Ding, Liying Cheng, Jia-Wei
  Low, Lidong Bing, and Luo Si. 2021.
\newblock On the effectiveness of adapter-based tuning for pretrained language
  model adaptation.
\newblock \emph{arXiv preprint arXiv:2106.03164}.

\bibitem[{Houlsby et~al.(2019)Houlsby, Giurgiu, Jastrzebski, Morrone,
  De~Laroussilhe, Gesmundo, Attariyan, and Gelly}]{houlsby2019parameter}
Neil Houlsby, Andrei Giurgiu, Stanislaw Jastrzebski, Bruna Morrone, Quentin
  De~Laroussilhe, Andrea Gesmundo, Mona Attariyan, and Sylvain Gelly. 2019.
\newblock Parameter-efficient transfer learning for nlp.
\newblock In \emph{International Conference on Machine Learning}, pages
  2790--2799. PMLR.

\bibitem[{Hu and Liu(2004)}]{hu2004mining}
Minqing Hu and Bing Liu. 2004.
\newblock Mining and summarizing customer reviews.
\newblock In \emph{Proceedings of the tenth ACM SIGKDD international conference
  on Knowledge discovery and data mining}, pages 168--177.

\bibitem[{Lester et~al.(2021)Lester, Al-Rfou, and Constant}]{lester2021power}
Brian Lester, Rami Al-Rfou, and Noah Constant. 2021.
\newblock The power of scale for parameter-efficient prompt tuning.
\newblock \emph{arXiv preprint arXiv:2104.08691}.

\bibitem[{Li et~al.(2020)Li, Zhou, He, Wang, Yang, and Li}]{li2020sentence}
Bohan Li, Hao Zhou, Junxian He, Mingxuan Wang, Yiming Yang, and Lei Li. 2020.
\newblock On the sentence embeddings from pre-trained language models.
\newblock \emph{arXiv preprint arXiv:2011.05864}.

\bibitem[{Li and Liang(2021)}]{li2021prefix}
Xiang~Lisa Li and Percy Liang. 2021.
\newblock Prefix-tuning: Optimizing continuous prompts for generation.
\newblock \emph{arXiv preprint arXiv:2101.00190}.

\bibitem[{Liu et~al.(2021{\natexlab{a}})Liu, Ji, Fu, Du, Yang, and
  Tang}]{liu2021p}
Xiao Liu, Kaixuan Ji, Yicheng Fu, Zhengxiao Du, Zhilin Yang, and Jie Tang.
  2021{\natexlab{a}}.
\newblock P-tuning v2: Prompt tuning can be comparable to fine-tuning
  universally across scales and tasks.
\newblock \emph{arXiv preprint arXiv:2110.07602}.

\bibitem[{Liu et~al.(2021{\natexlab{b}})Liu, Zheng, Du, Ding, Qian, Yang, and
  Tang}]{liu2021gpt}
Xiao Liu, Yanan Zheng, Zhengxiao Du, Ming Ding, Yujie Qian, Zhilin Yang, and
  Jie Tang. 2021{\natexlab{b}}.
\newblock Gpt understands, too.
\newblock \emph{arXiv preprint arXiv:2103.10385}.

\bibitem[{Liu et~al.(2019)Liu, Ott, Goyal, Du, Joshi, Chen, Levy, Lewis,
  Zettlemoyer, and Stoyanov}]{liu2019roberta}
Yinhan Liu, Myle Ott, Naman Goyal, Jingfei Du, Mandar Joshi, Danqi Chen, Omer
  Levy, Mike Lewis, Luke Zettlemoyer, and Veselin Stoyanov. 2019.
\newblock Roberta: A robustly optimized bert pretraining approach.
\newblock \emph{arXiv preprint arXiv:1907.11692}.

\bibitem[{Mahabadi et~al.(2021)Mahabadi, Henderson, and
  Ruder}]{mahabadi2021compacter}
Rabeeh~Karimi Mahabadi, James Henderson, and Sebastian Ruder. 2021.
\newblock Compacter: Efficient low-rank hypercomplex adapter layers.
\newblock \emph{arXiv preprint arXiv:2106.04647}.

\bibitem[{Ott et~al.(2019)Ott, Edunov, Baevski, Fan, Gross, Ng, Grangier, and
  Auli}]{ott2019fairseq}
Myle Ott, Sergey Edunov, Alexei Baevski, Angela Fan, Sam Gross, Nathan Ng,
  David Grangier, and Michael Auli. 2019.
\newblock fairseq: A fast, extensible toolkit for sequence modeling.
\newblock In \emph{Proceedings of NAACL-HLT 2019: Demonstrations}.

\bibitem[{Pang and Lee(2004)}]{pang2004sentimental}
Bo~Pang and Lillian Lee. 2004.
\newblock A sentimental education: Sentiment analysis using subjectivity
  summarization based on minimum cuts.
\newblock \emph{arXiv preprint cs/0409058}.

\bibitem[{Pang and Lee(2005)}]{pang2005seeing}
Bo~Pang and Lillian Lee. 2005.
\newblock Seeing stars: Exploiting class relationships for sentiment
  categorization with respect to rating scales.
\newblock \emph{arXiv preprint cs/0506075}.

\bibitem[{Pennington et~al.(2014)Pennington, Socher, and
  Manning}]{pennington2014glove}
Jeffrey Pennington, Richard Socher, and Christopher~D Manning. 2014.
\newblock Glove: Global vectors for word representation.
\newblock In \emph{Proceedings of the 2014 conference on empirical methods in
  natural language processing (EMNLP)}, pages 1532--1543.

\bibitem[{Pfeiffer et~al.(2020{\natexlab{a}})Pfeiffer, Kamath, R{\"u}ckl{\'e},
  Cho, and Gurevych}]{pfeiffer2020adapterfusion}
Jonas Pfeiffer, Aishwarya Kamath, Andreas R{\"u}ckl{\'e}, Kyunghyun Cho, and
  Iryna Gurevych. 2020{\natexlab{a}}.
\newblock Adapterfusion: Non-destructive task composition for transfer
  learning.
\newblock \emph{arXiv preprint arXiv:2005.00247}.

\bibitem[{Pfeiffer et~al.(2020{\natexlab{b}})Pfeiffer, Vuli{\'c}, Gurevych, and
  Ruder}]{pfeiffer2020mad}
Jonas Pfeiffer, Ivan Vuli{\'c}, Iryna Gurevych, and Sebastian Ruder.
  2020{\natexlab{b}}.
\newblock Mad-x: An adapter-based framework for multi-task cross-lingual
  transfer.
\newblock \emph{arXiv preprint arXiv:2005.00052}.

\bibitem[{Phang et~al.(2018)Phang, F{\'e}vry, and Bowman}]{phang2018sentence}
Jason Phang, Thibault F{\'e}vry, and Samuel~R Bowman. 2018.
\newblock Sentence encoders on stilts: Supplementary training on intermediate
  labeled-data tasks.
\newblock \emph{arXiv preprint arXiv:1811.01088}.

\bibitem[{Reimers and Gurevych(2019)}]{reimers2019sentence}
Nils Reimers and Iryna Gurevych. 2019.
\newblock Sentence-bert: Sentence embeddings using siamese bert-networks.
\newblock \emph{arXiv preprint arXiv:1908.10084}.

\bibitem[{Schick and Sch{\"u}tze(2020)}]{schick2020exploiting}
Timo Schick and Hinrich Sch{\"u}tze. 2020.
\newblock Exploiting cloze questions for few shot text classification and
  natural language inference.
\newblock \emph{arXiv preprint arXiv:2001.07676}.

\bibitem[{Shin et~al.(2020)Shin, Razeghi, Logan~IV, Wallace, and
  Singh}]{shin2020autoprompt}
Taylor Shin, Yasaman Razeghi, Robert~L Logan~IV, Eric Wallace, and Sameer
  Singh. 2020.
\newblock Autoprompt: Eliciting knowledge from language models with
  automatically generated prompts.
\newblock \emph{arXiv preprint arXiv:2010.15980}.

\bibitem[{Wang et~al.(2018)Wang, Singh, Michael, Hill, Levy, and
  Bowman}]{wang2018glue}
Alex Wang, Amanpreet Singh, Julian Michael, Felix Hill, Omer Levy, and Samuel~R
  Bowman. 2018.
\newblock Glue: A multi-task benchmark and analysis platform for natural
  language understanding.
\newblock \emph{arXiv preprint arXiv:1804.07461}.

\bibitem[{Wang et~al.(2021)Wang, Fang, Khabsa, Mao, and
  Ma}]{wang2021entailment}
Sinong Wang, Han Fang, Madian Khabsa, Hanzi Mao, and Hao Ma. 2021.
\newblock Entailment as few-shot learner.
\newblock \emph{arXiv preprint arXiv:2104.14690}.

\bibitem[{Wiebe et~al.(2005)Wiebe, Wilson, and Cardie}]{wiebe2005annotating}
Janyce Wiebe, Theresa Wilson, and Claire Cardie. 2005.
\newblock Annotating expressions of opinions and emotions in language.
\newblock \emph{Language resources and evaluation}, 39(2):165--210.

\bibitem[{Zhang et~al.(2021)Zhang, Tay, Zhang, Chan, Luu, Hui, and
  Fu}]{zhang2021beyond}
Aston Zhang, Yi~Tay, Shuai Zhang, Alvin Chan, Anh~Tuan Luu, Siu~Cheung Hui, and
  Jie Fu. 2021.
\newblock Beyond fully-connected layers with quaternions: Parameterization of
  hypercomplex multiplications with $1/n $ parameters.
\newblock \emph{arXiv preprint arXiv:2102.08597}.

\end{thebibliography}
\bibliographystyle{main_NAACL2022}

\clearpage

\appendix
\onecolumn
\section{Appendix} \label{sec:appendix}
\subsection{Experimental Settings}\label{sec:aexp}
\subsubsection{Training hyperparameters}
We use RoBERTa-Large~\cite{liu2019roberta} model implemented by Fairseq~\cite{ott2019fairseq} as our basic model. The detailed model hyperparameters are listed below:

\begin{table}[h]
\footnotesize
\centering
\begin{adjustbox}{max width=\textwidth}
\begin{tabular}{lccc}
    \toprule
    \textbf{Hyperparam} & \textbf{Supplmentary} & \textbf{Finetune} & \textbf{few-shot} \\
    \midrule
 \#Layers &  24 & 24 & 24\\ 
      \midrule
 Hidden size & 1024 & 1024 & 1024\\
     \midrule
 FFN inner hidden size & 4096 & 4096 & 4096 \\
      \midrule
 Attention heads  & 16  & 16 & 16\\
    \midrule
 Attention head size  & 64 & 64 & 64\\
 \midrule
 dropout  & 0.1  & 0.1 & 0.1 \\
 \midrule
 Learning Rate & linearly decayed & fixed & fixed\\
 \midrule
 Peak Learning Rate & $1e^{-5}$ & \{$5e^{-3}, 1e^{-3}, 5e^{-4}, 1e^{-4}$\} & $5e^{-4}$ \\
 \midrule
 Batch Size & 32 & \{16, 32\} & 16 \\
 \midrule
 Weight Decay & 0.1 & 0.1 & 0.1\\
 \midrule
 Training Epoch & 10 & 50 & 50\\
 \midrule
 Adam $\epsilon$ & $1e^{-6}$ & $1e^{-6}$ & $1e^{-6}$ \\
 \midrule
 Adam $\beta_1$ & 0.9 & 0.9 & 0.9 \\
 \midrule
 Adam $\beta_2$ & 0.98 & 0.98 & 0.98\\
\bottomrule
\end{tabular}
\end{adjustbox}
\caption{
    Hyperparameters for supplmentary training, fine-tuning, few-shot fine-tuning.
}
\label{tab:exp-detail}
\end{table}

Note that for both transformer fine-tuning methods including RoBERTa~\citep{liu2019roberta} and EFL~\citep{wang2021entailment}, we follow their official training instructions, i.e., using a polynomial learning rate scheduler with 6\% of total steps to warm up and tuning for 10 epochs.

\subsubsection{Model hyperparameters}
We report the detailed model hyperparameters for each method in Table~\ref{tab:main} and illustrate how numbers in Table~\ref{tab:efficiency} are computed. 

\textbf{Compacter:} hidden size $d = 1024$, adapter hidden size $m=16$, user defined $n=4$, each transformer layer inserts 2 compacters. Down-project $s_i$ matrix takes $1024/4 \times 4 \times 24 \times 2 = 48K$, down-project $t_i$ matrix takes $16/4 \times 4 \times 24 \times 2 = 0.75K$, hidden bias takes $16 \times 24 \times 2 = 0.75K$, up-project $s_i$ and $t_i$ matrix takes the same number of parameters as down-projector, the output bias takes $1024 \times 24 \times 2 = 48K$, the shared matrix $A_i$ takes $4^3 \times 24 \times 2 = 3K$. 
\textbf{Total parameters}:~$48 + 0.75 + 0.75 + 48 + 0.75 + 48 + 3 =149.25K$. 

\textbf{Adapter:} hidden size $d = 1024$, adapter hidden size $m=16$.
\textbf{Total parameters}:~$(1024 \times 16 + 16 + 16 \times 1024 + 1024) \times 24 \times 2 = 1.55M$.

\textbf{Prompt-tuning:} prompt length $t=5$.
\textbf{Total parameters}:~$5 \times 1024 = 5K$.

\textbf{Prompt-tuning-134:} prompt length $t=134$.
\textbf{Total parameters}:~$134 \times 1024 = 134K$.

\textbf{P-tuning v2:} prompt length $t=5$, inserted layers $24$.
\textbf{Total parameters}:~$5 \times 24 \times 1024 = 120K$.

\textbf{S-IDPG-PHM:} hidden size $d = 1024$, generator hidden size $m=256$, prompt length $t=5$, user defined $n=16$ (Cf. Equation~\ref{eq4}). First PHM layer \textbf{$W_1$} takes $1024 / 16 \times 256 / 16 \times 16 + 256 = 16.25K$ parameters, second PHM layer \textbf{$W_2$} takes $256/16 \times 5 \times 1024 /16 \times 16 + 5 \times 1024 = 85K$ parameters, the shared matrix $A_i$ takes $16^3 = 4K$ (Note we use one shared matrix in single version IDPG).
\textbf{Total parameters}:~$105K$.

\textbf{S-IDPG-DNN:} hidden size $d = 1024$, generator hidden size $m=256$, prompt length $t=5$. 
\textbf{Total parameters}:~$1024 \times 256 + 256 + 256 \times 5 \times 1024 + 5 \times 1024 = 1.5M$.

\textbf{M-IDPG-PHM-GloVe:} input vector size $300$, generator hidden size $m=16$, prompt length $t=5$, user defined $n=4$ (Cf. Equation~\ref{eq4}). First PHM layer \textbf{$W_1$} takes $300 / 4 \times 16 / 4 \times 4 + 16 = 1216$ parameters, second PHM layer \textbf{$W_2$} takes $16/4 \times 5 \times 1024 /4 \times 4 + 5 \times 1024 \times 24 = 140K$ parameters, the shared matrix $A_i$ takes $4^3 \times 2 = 128$. 
\textbf{Total parameters}:~$141K$.

\textbf{M-IDPG-PHM:} hidden size $d = 1024$, generator hidden size $m=16$, prompt length $t=5$, user defined $n=16$ (Cf. Equation~\ref{eq4}). First PHM layer \textbf{$W_1$} takes $1024 / 16 \times 16 / 16 \times 16 + 16 = 1K$ parameters, second PHM layer \textbf{$W_2$} takes $16/16 \times 5 \times 1024 /16 \times 16 + 5 \times 1024 \times 24 = 125K$ parameters, the shared matrix $A_i$ takes $16^3 16 \times 2 = 8K$. 
\textbf{Total parameters}:~$134K$.

\textbf{M-IDPG-DNN:} hidden size $d = 1024$, generator hidden size $m=16$, prompt length $t=5$. 
\textbf{Total parameters}:~$1024 \times 16 + 16 + 16 \times 5 \times 1024 + 5 \times 1024 \times 24 = 216K$.

\subsection{Datasets}
We provide a detailed information in Table~\ref{tab:datasets} for 10 NLU datasets we used. 
\begin{table}[h]
\footnotesize
\centering
\resizebox{0.8\columnwidth}{!}{
\begin{tabular}{lcccc}
    \toprule
    \textbf{Corpus} & |\textbf{Train}| & |\textbf{Valiadation}| & \textbf{Task} & \textbf{Evaluation Metrics} \\
    \midrule
    Single Sentence Tasks & \\
    \midrule
    CR &  1,775 & 2,000 & sentiment & accuracy \\ 
    MR & 8,662 & 2,000 & sentiment & accuracy \\
    SUBJ & 8,000 & 2,000 & sentiment & accuracy \\
    MPQA & 8,606 & 2,000 & opinion polarity & accuracy \\
 SST-2 & 67,349 & 1,821 & sentiment analysis & accuracy \\ 
 \midrule
 Sentence Pair Tasks & \\
 \midrule
    QNLI & 104,743 & 5,463 & NLI & accuracy \\
    RTE & 2,491 & 278 & NLI & accuracy \\
    MRPC & 3,668 & 409 & paraphrase & accuracy/F1 \\
    QQP & 363,846 & 40,430 & paraphrase & accuracy/F1 \\
    STS-B & 5,749 & 1,500 &  sentence similarity & Pearson/Spearman corr.\\
\bottomrule
\end{tabular}
}
\caption{
    The datasets evaluated in this work.
}
\label{tab:datasets}
\end{table}

\begin{table*}[t]
\caption{Main results of different transfer learning method. Each methods are evaluated on full test sets (dev sets for GLUE tasks).
We report average results across 5 runs with different initialization. We report the average of accuracy and F1 for both MRPC and QQP, and average of Pearson and Spearman correlation coefficients for STS-B. For all the other tasks, we report accuracy.} 
\label{tab:midpg}
\centering
\setlength{\tabcolsep}{8pt}
\begin{adjustbox}{max width=1.0\textwidth}
\begin{tabular}{lc|cccccccccc|c}  %
\toprule
 \textbf{Method} & \textbf{m} & \textbf{MPQA} & \textbf{Subj} & \textbf{CR} & \textbf{MR} & \textbf{SST-2} & \textbf{QNLI} & \textbf{RTE} & \textbf{MRPC} & \textbf{STS-B} & \textbf{QQP} & Avg\\ 
\midrule
\multicolumn{6}{l}{\emph{Input: Layer 0}} \\
\midrule
S version & 256 & 91.2$_{\pm0.2}$ & 97.6$_{\pm0.1}$ & 93.8$_{\pm 0.3}$ & 92.6$_{\pm0.2}$ & 95.9$_{\pm0.1}$ & {93.8}$_{\pm 0.1}$ & 79.9$_{\pm8.0}$ & {90.8}$_{\pm0.5}$ & 90.9$_{\pm0.4}$ & {85.9}$_{\pm0.4}$ & 91.2 \\
M version & 256 & 91.2$_{\pm0.3}$ & 97.5$_{\pm0.1}$ & 93.6$_{\pm 0.3}$ & 92.7$_{\pm0.2}$ & {95.7}$_{\pm0.2}$ & 94.3$_{\pm 0.1}$ & 85.5$_{\pm1.0}$ & 91.8$_{\pm0.3}$ & {91.4}$_{\pm0.2}$ & {87.0}$_{\pm0.4}$ & 92.1 \\
L version & 256 & 91.3$_{\pm0.1}$ & 97.6$_{\pm0.2}$ & 93.8$_{\pm 0.3}$ & 92.6$_{\pm0.1}$ & {95.5}$_{\pm0.2}$ & 94.5$_{\pm 0.2}$ & 86.5$_{\pm0.5}$ & 92.5$_{\pm0.8}$ & {91.6}$_{\pm0.1}$ & {87.3}$_{\pm0.3}$ & 92.3 \\
\midrule
\multicolumn{6}{l}{\emph{Input: Previous Layer}} \\
\midrule
S version & 256 & 91.2$_{\pm0.2}$ & 97.5$_{\pm0.1}$ & 93.5$_{\pm 0.3}$ & 92.6$_{\pm0.1}$ & 95.8$_{\pm0.3}$ & {94.0}$_{\pm 0.1}$ & 83.4$_{\pm1.5}$ & {91.9}$_{\pm0.3}$ & 91.1$_{\pm0.3}$ & {86.9}$_{\pm0.2}$ & 91.8 \\
M version & 256 & 91.0$_{\pm0.2}$ & 97.5$_{\pm0.1}$ & 93.4$_{\pm 0.4}$ & 92.6$_{\pm0.2}$ & {96.0}$_{\pm0.1}$ & 94.4$_{\pm 0.2}$ & 86.6$_{\pm1.2}$ & 91.5$_{\pm0.4}$ & {91.4}$_{\pm0.2}$ & {86.3}$_{\pm0.1}$ & 92.1 \\
L version & 256 & 91.3$_{\pm0.2}$ & 97.4$_{\pm0.0}$ & 93.3$_{\pm 0.3}$ & 92.5$_{\pm0.2}$ & {95.8}$_{\pm0.1}$ & 94.5$_{\pm 0.3}$ & 86.9$_{\pm0.8}$ & 92.1$_{\pm0.4}$ & {91.7}$_{\pm0.2}$ & {87.1}$_{\pm0.2}$ & 92.3 \\
\midrule
 \multicolumn{6}{l}{\emph{Input: Previous Layer}} \\
\midrule
S version & 16 & 91.4$_{\pm0.2}$ & 97.5$_{\pm0.1}$ & 93.6$_{\pm 0.2}$ & 92.5$_{\pm0.2}$ & 95.7$_{\pm0.2}$ & {93.9}$_{\pm 0.0}$ & 83.6$_{\pm0.8}$ & {91.9}$_{\pm0.4}$ & 90.9$_{\pm0.3}$ & {85.5}$_{\pm0.4}$ & 91.6 \\
M version & 16 & 91.2$_{\pm0.2}$ & 97.5$_{\pm0.1}$ & 93.4$_{\pm 0.3}$ & 92.6$_{\pm0.3}$ & {96.0}$_{\pm0.3}$ & 94.5$_{\pm 0.1}$ & 83.5$_{\pm0.7}$ & 92.3$_{\pm0.2}$ & {91.4}$_{\pm0.4}$ & {87.1}$_{\pm0.1}$ & 92.0 \\

\bottomrule
\end{tabular}
\end{adjustbox}
\end{table*}
\subsection{Detailed results for Multi-layer Architecture Exploration}\label{sec:A3}
We provide a detailed result table for all compared methods in Section~\ref{4-5-3}. Note that the M version model with $m=16$ and previous layer as input one is slightly higher than the results shown in Table~\ref{tab:main}(Cf. M-IDPG-PHM), this is because we tune the learning rate more carefully in Table~\ref{tab:midpg} ($lr \in \{1e^{-2}, 7e^{-3}, 5e^{-3}, 3e^{-3}, 1e^{-3}, 7e^{-4}, 5e^{-4}, 3e^{-4}, 1e^{-4}\}$) to seek the best performance each model can reach. While in Table~\ref{tab:main}, we tune the learning rate from $\{5e^{-3}, 1e^{-3}, 5e^{-4}, 1e^{-4}\}$ to make the fair comparison with other models. 

\begin{figure}[htbp]
	\centering
	\begin{center}
		\includegraphics[width=0.45\textwidth]{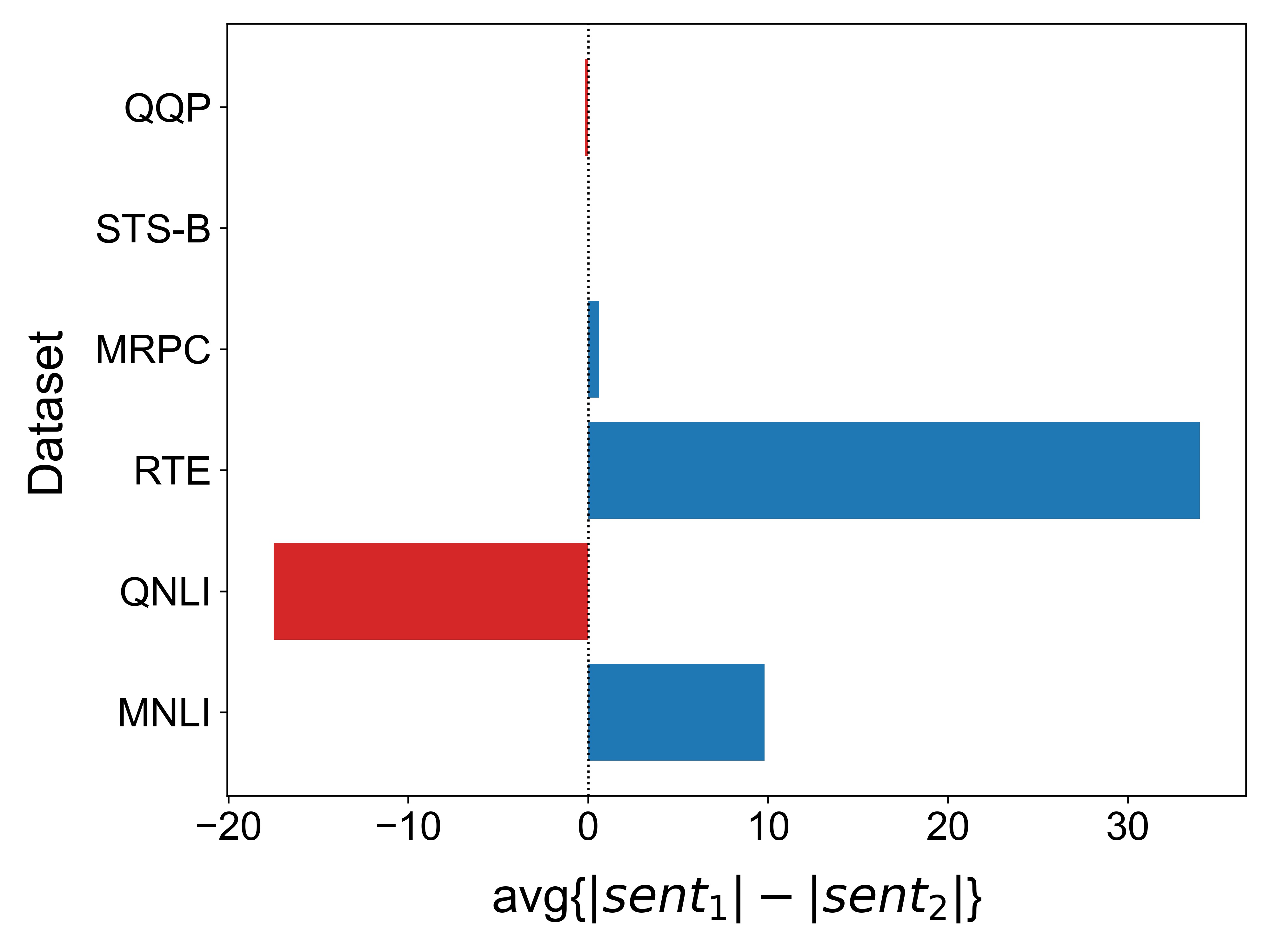}
		\caption[7.5pt]{Length difference of GLUE sentence pair datasets.}
		\vspace{-0.25in}
		\label{fig:dis-of-sp}
	\end{center}
\end{figure}
\subsection{Supplementary Training for Single-layer IDPG} \label{sec-sup}
According to previous works~\cite{phang2018sentence, wang2021entailment}, supplementing pre-trained LMs with rich data helps tasks with limited labels and stabilizes downstream fine-tuning. Following this idea, we conduct intermediate training for single-layer IDPG.

\begin{figure}[htbp]
	\centering
	\begin{center}
	\vspace{-0.1in}
		\includegraphics[width=0.49\textwidth]{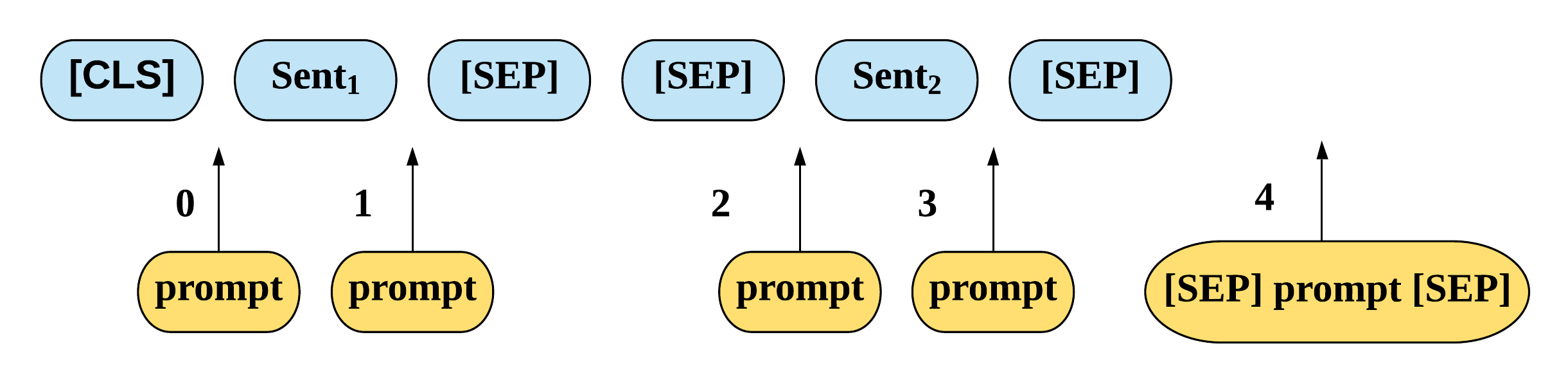}
		\caption[7.5pt]{Insertion positions for sentence-pair tasks.}
		\vspace{-0.2in}
		\label{fig:pos}
	\end{center}
\end{figure}

However, a drawback of supplementary training is that if the data distribution of the downstream tasks is quite different from the supplementary training task, i.e., MRPC vs.~MNLI~\cite{wang2018glue}, it may harm the downstream performance. Figure~\ref{fig:dis-of-sp} provides a comprehensive statistic among all sentence pair tasks in GLUE benchmark. For example, the length of the first sentence in MNLI is $9.8$ longer than the second sentence on average, while this length difference in MRPC is only $0.6$.
One natural solution to smooth the length distribution difference between tasks is to insert prompt in both supplementary training and downstream fine-tuning stage. For example, assuming that we are adding a prompt with a length $t=5$ after the second sentence in the supplementary training stage on MNLI. Then, when fine-tuning downstream tasks such as MRPC, we concatenate the prompt after the first sentence. In this way, the length difference in MNLI and MRPC becomes more balanced: $4.8$ vs.~$0.6 + 5 = 5.6$. As shown in Figure~\ref{fig:pos}, we test five different insertion positions (Pos 0--4) for sentence pair tasks and three different positions (Pos 0, 1, 4) for single sentence tasks.
We further reduce the distribution difference by reconstructing the supplementary training data. We double the MNLI dataset by reordering the two sentences on one shard, and use the doubled dataset during intermediate training.

\begin{table}[h]
\footnotesize
\centering
\resizebox{0.5\columnwidth}{!}{
\begin{tabular}{lcc}
    \toprule
    \textbf{Architecture} & \textbf{Avg} & \textbf{Voting}  \\
    \midrule
 PHM &  86.1 & 86.9 \\ 
      \midrule
 +residual & 85.9 & 86.7 \\
     \midrule
 +LayerNorm &  86.1 & 87.1  \\
      \midrule
 +residual+LayerNorm  & 77.8  & 81.2 \\
    \bottomrule
\end{tabular}
}
\caption{
    Ablation study on generator architecture. 
}
\vspace{-0.15in}
\label{tab:abla-arch}
\end{table}

\begin{figure}[t]
	\centering
	\begin{center}
		\includegraphics[width=0.8\textwidth]{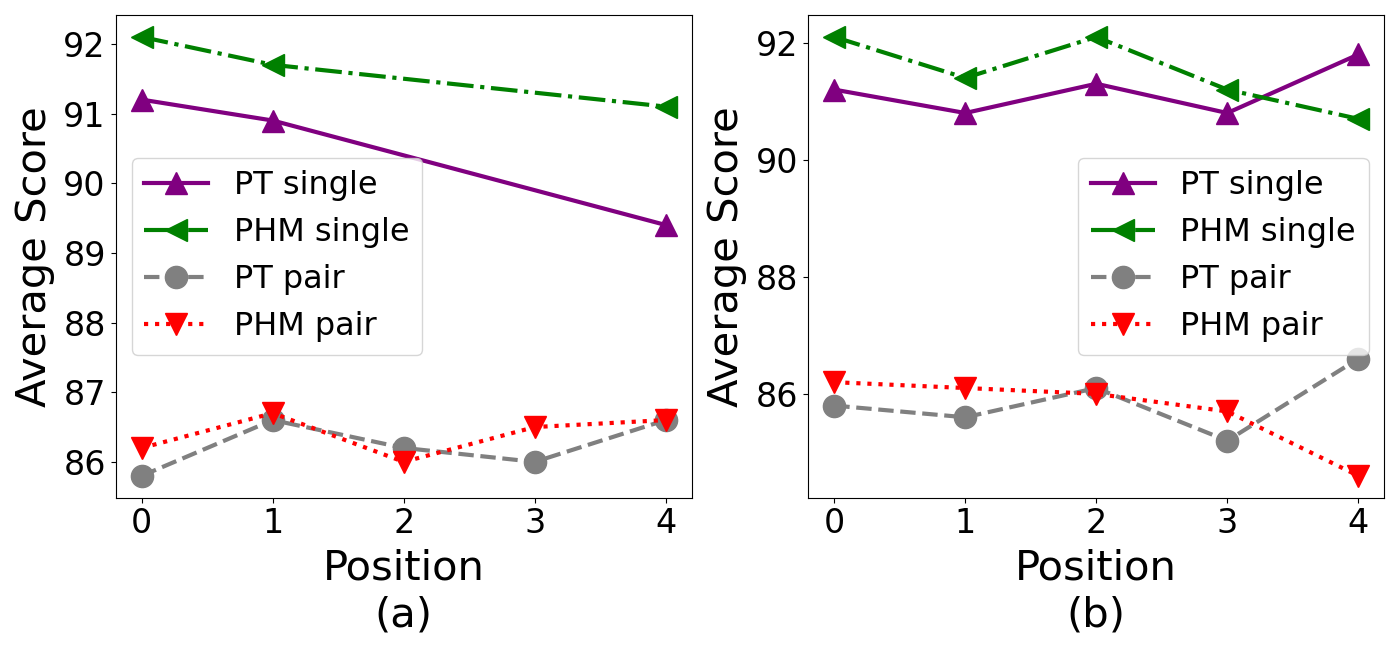}
		\caption[7.5pt]{Impact of prompt position on (a)~downstream tasks; (b)~supplementary training phase.}
		\label{fig:ablapos}
		\vspace{-0.2in}
	\end{center}
\end{figure}
\subsection{Ablation study for single-layer IDPG}\label{appendix-ablation}
\subsubsection{Generator Architecture Exploration}
We explore three different architectures for the proposed PHM-based generator: (i) Residual: a residual structure~\cite{he2016deep} is applied to add the sentence representation to each generated tokens; (ii) LayerNorm: layer normalization~\cite{ba2016layer} is also added to normalize the generated token embedding; (iii) residual + layerNorm: a mixed model that uses both the residual component and LayerNorm. Note that, to balance the token embedding and sentence embedding, we apply LayerNorm to each embedding first, then after the add-up, use LayerNorm again to control the generated tokens.
We observe that adding LayerNorm slightly improves the voting results, while residual performs slightly worse. One surprising result is that the mixed model of Residual and LayerNorm has significantly poorer performance compared to other methods.

\subsubsection{Prompt Position}
As we discussed in Section~\ref{sec-sup}, the prompt position has a direct impact on the prediction results. We conduct a comprehensive study of the prompt position for our proposed method in both supplementary training and downstream fine-tuning phases.

Looking at the prompt position in downstream tasks first, Figure~\ref{fig:ablapos}(a) shows that for both standard prompt tuning and our proposed method, the best position is 0 for single-sentence tasks and 1 for sentence-pair tasks. This result is intuitive for single-sentence tasks since prompt in position 0 can be regarded as the premise and original input sentence as the hypothesis. For sentence-pair tasks, we hypothesize that inserting prompt into position 1 can better align the two input sentences. Figure~\ref{fig:ablapos}(b) illustrates the effect of prompt position on the supplementary training phase. It is interesting that IDPG achieves best results in position 0 while the standard prompt-tuning achieves the best results in position 4 for both single-sentence and sentence-pair tasks.

\subsection{Cosine Similarity Distributions in STS-B}\label{appendix-cos}
\begin{figure}[t]
	\centering
\begin{subfigure}{0.45\linewidth}
\centering
\includegraphics[width=0.9\textwidth]{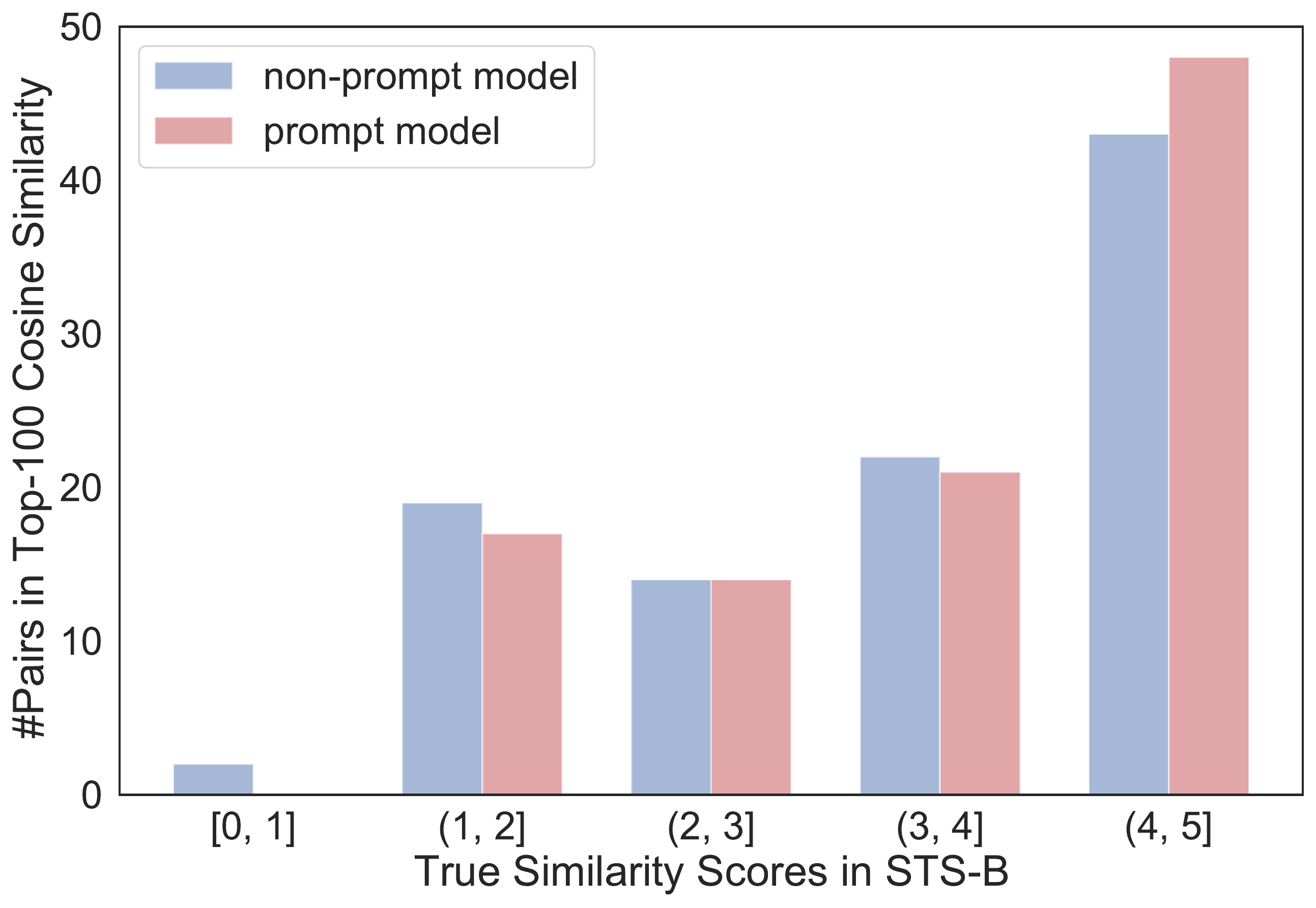}
		\caption{The number of pairs of each group in Top-100 cosine similarity ranking. }
		\label{fig:ana100}
\end{subfigure}
\hfil
\begin{subfigure}{0.45\linewidth}
\centering
\includegraphics[width=0.9\textwidth]{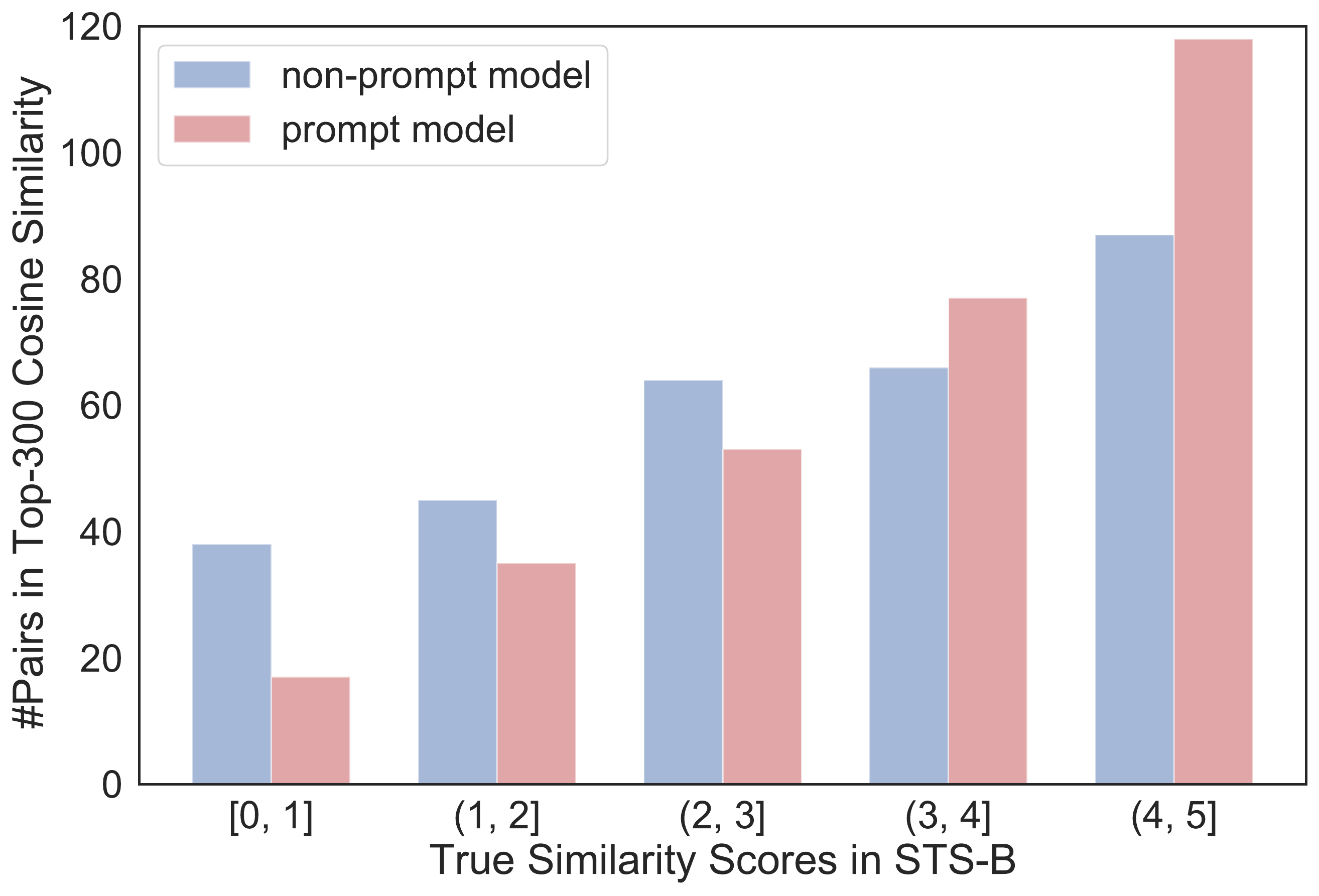}
		\caption{The number of pairs of each group in Top-300 cosine similarity ranking.}
		\label{fig:ana300}
\end{subfigure}
\caption[7.5pt]{The number of pairs of each group in Top-k cosine similarity ranking.}\label{fig:abl-ana-full}
\end{figure}

We present the cosine similarity distributions when $k=100$ and $k=300$ in Figure~\ref{fig:ana100} and in Figure~\ref{fig:ana300}, respectively. 

\subsection{Ablation Study on Prompt Length}\label{appendix-len}
\begin{figure}[t]
	\centering
	\begin{center}
		\includegraphics[width=0.5\textwidth]{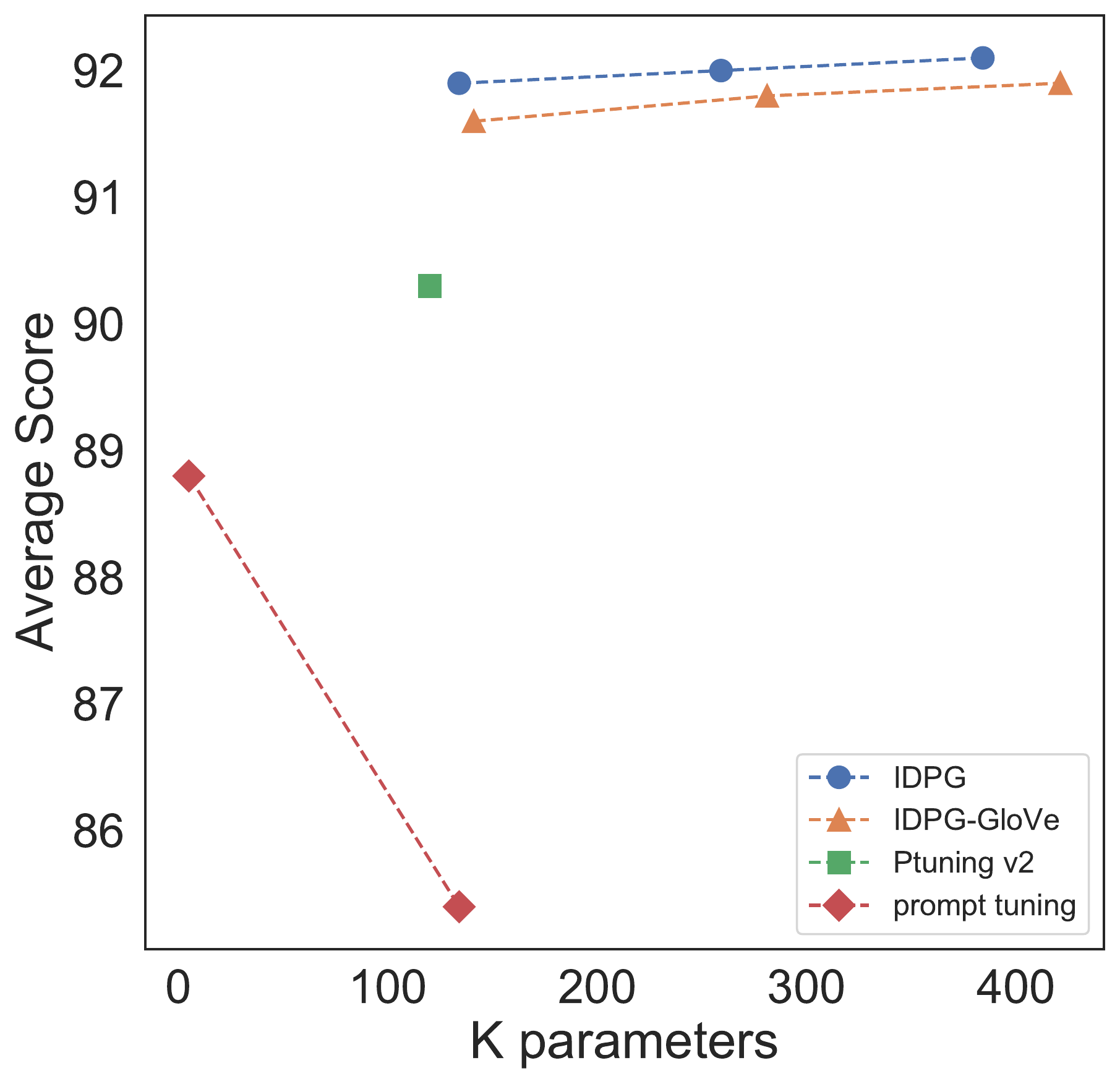}
		\caption[7.5pt]{Impact of prompt length.}
		\label{fig:abl-len}
		\vspace{-0.2in}
	\end{center}
\end{figure}

We present the impact of prompt length among several prompt tuning methods in Figure~\ref{fig:abl-len}. IDPG shows its stability when scaling to larger models with longer prompts. 

\subsection{Potential Risks}
Our proposed model IDPG is a novel efficient transfer learning method. It tunes small portion parameters while directly employs backbone model parameters without any changing. 
However, if the backbone model stored online is attacked, whether IDPG could still work well remains unknown. One should be careful to apply our proposed model and all other prompt tuning methods in high-stakes areas without a comprehensive test. 

\end{document}